\RequirePackage{amsthm}
\documentclass[pdflatex,sn-basic]{sn-jnl}

\usepackage{graphicx}%
\usepackage{multirow} %
\usepackage{amsmath,amssymb,amsfonts}%
\usepackage{amsthm}%
\usepackage{mathrsfs}%
\usepackage[title]{appendix}%
\usepackage{textcomp}%
\usepackage{manyfoot}%
\usepackage{booktabs}%
\usepackage{algorithm}%
\usepackage{algorithmicx}%
\usepackage{algpseudocode}%
\usepackage{listings}%

\usepackage{balance}
\usepackage{hyperref}
\usepackage{rotating}
\usepackage{makecell}
\usepackage{subfig}

\usepackage{soul}
\usepackage{comment}
\usepackage{tabulary}
\usepackage{array}
\usepackage{threeparttable}
\usepackage{natbib}
\usepackage[table]{xcolor}

\theoremstyle{thmstyleone}%
\theoremstyle{thmstyletwo}%

\theoremstyle{thmstylethree}%

\begin{document}

\title[Safeguarding Privacy: Privacy-Preserving Detection of Mind Wandering and Disengagement Using Federated Learning in Online Education]{Safeguarding Privacy: Privacy-Preserving Detection of Mind Wandering and Disengagement Using Federated Learning in Online Education\\ 
\textit{\small A Preprint}}


\author[1]{\fnm{Anna} \sur{Bodonhelyi}}\email{anna.bodonhelyi@tum.de}
\equalcont{These authors contributed equally to this work.}

\author[1]{\fnm{Mengdi} \sur{Wang}}\email{mengdi.wang@tum.de}
\equalcont{These authors contributed equally to this work.}

\author[1]{\fnm{Efe} \sur{Bozkir}}\email{efe.bozkir@tum.de}

\author[1]{\fnm{Babette} \sur{Bühler}}\email{babette.bühler@tum.de}

\author[1]{\fnm{Enkelejda} \sur{Kasneci}}\email{enkelejda.kasneci@tum.de}

\affil[1]{\orgdiv{Chair of Human-Centered Technologies for Learning}, \orgname{Technical University of Munich}, \orgaddress{\street{Arcisstr. 21.}, \city{Munich}, \postcode{80333}, \state{Bayern}, \country{Germany}}}


\abstract{Since the COVID-19 pandemic, online courses have expanded access to education, yet the absence of direct instructor support challenges learners' ability to self-regulate attention and engagement. Mind wandering and disengagement can be detrimental to learning outcomes, making their automated detection via video-based indicators a promising approach for real-time learner support. However, machine learning-based approaches often require sharing sensitive data, raising privacy concerns. Federated learning offers a privacy-preserving alternative by enabling decentralized model training while also distributing computational load. We propose a framework exploiting cross-device federated learning to address different manifestations of behavioral and cognitive disengagement during remote learning, specifically behavioral disengagement, mind wandering, and boredom. We fit video-based cognitive disengagement detection models using facial expressions and gaze features. By adopting federated learning, we safeguard users' data privacy through privacy-by-design and introduce a novel solution with the potential for real-time learner support. We further address challenges posed by eyeglasses by incorporating related features, enhancing overall model performance. To validate the performance of our approach, we conduct extensive experiments on five datasets and benchmark multiple federated learning algorithms. Our results show great promise for privacy-preserving educational technologies promoting learner engagement.}




\maketitle

\insert\footins{\noindent\footnotesize Preprint submitted to the International Journal of Artificial Intelligence in Education}

\begin{figure}
    \centering
    \includegraphics[width=\textwidth, keepaspectratio]{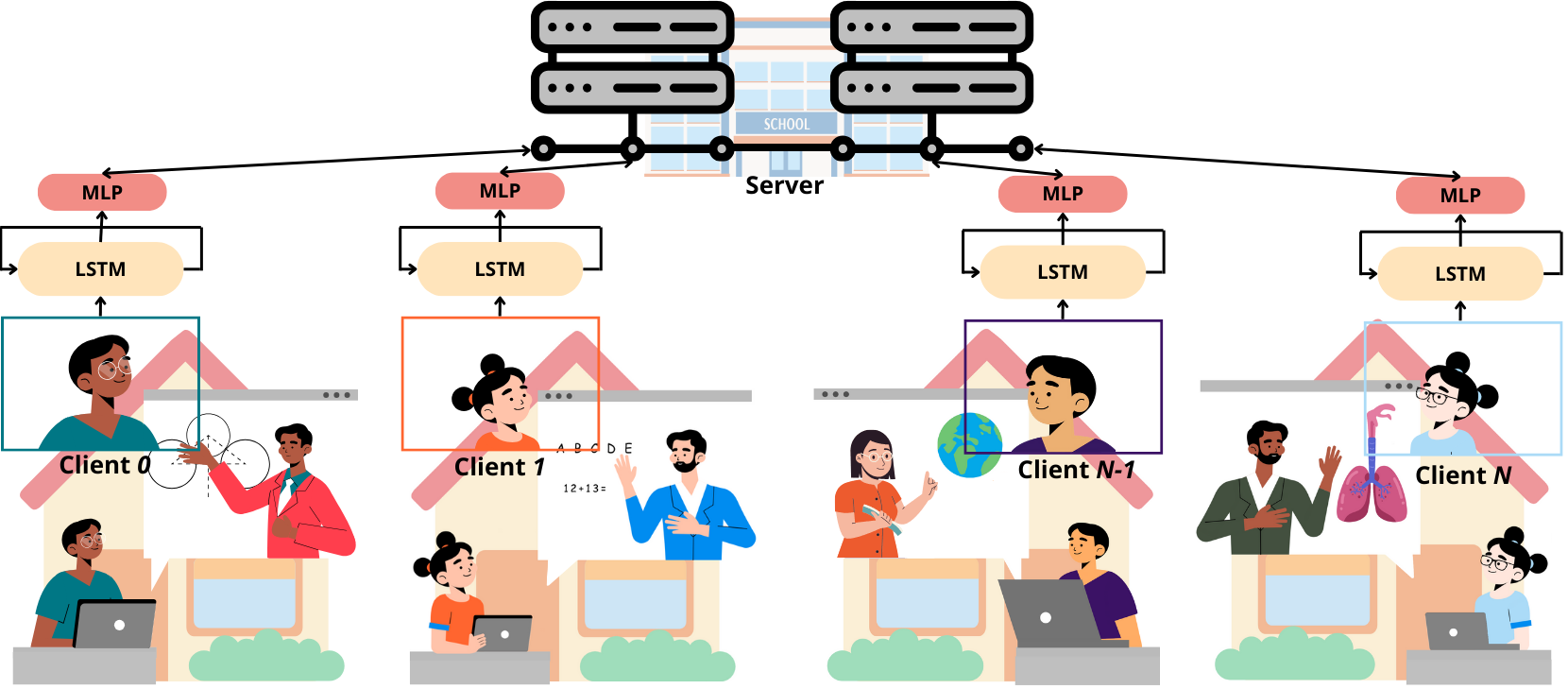}
    \captionof{figure}{Our proposed federated learning algorithm in an online learning scenario, where the server aggregates the trained client models on the extracted facial features from the students' video data to predict the remote learner state.}
\end{figure}

\section{Introduction}\label{chap:introduction}
Since the pandemic in 2019, an increasing shift of educational environments towards remote setup can be observed~\cite[][]{el2021did, simamora2020challenges, bylieva2021self}. In such a setup, students typically attend online lectures or watch pre-recorded videos using different devices such as tablets, laptops, and smartphones~\cite[][]{bodonhelyi2025modeling, bodonhelyi2025passive}. Whilst online learning shows a good potential to counteract physical and temporal barriers, conversely, in the absence of a guiding teacher in a remote learning scenario, it becomes more challenging for students to self-regulate for sustained focus~\cite[][]{wammes2017examining, buhler2024temporal, buhler2024task}. Additionally, distractions such as social media use are more common in remote learning~\cite[][]{hollis2016mind, wammes2019disengagement}. Behavioral and cognitive engagement play a central role in learning outcomes~\cite[][]{fredricks2004school,sumer2021multimodal,goldberg2021attentive}. Engagement can be conceptualized along three components: behavioral, cognitive, emotional, and agentic engagement~\cite[][]{fredricks2004school, sinatra2015challenges}. Behavioral disengagement during online learning, in the form of off-task behaviors, like checking emails or chatting, can negatively impact the learning outcome~\cite[][]{gobel2008student}. Another phenomenon, which is a form of covert cognitive disengagement and especially prevalent during online learning~\cite[][]{wammes2017examining}, is mind wandering, which describes the shift of attention to task-unrelated thought~\cite[][]{smallwood2006restless}. Mind wandering has been consistently related to poorer learning outcomes across tasks and age groups~\cite[][]{wong2022task}. Another prevalent learner state, that is a potential indicator of unsuccessful attentional engagement, is boredom~\cite[][]{westgate2018boring, eastwood2012unengaged}. Similar to disengagement and mind wandering, boredom is negatively related to academic achievement~\cite[][]{tze2016evaluating}. 
Therefore, detecting and mitigating learner states that hinder learning is crucial for educators in order to support learners adequately. 

In recent years, machine learning-driven automatic detection of off-task behaviors provides a promising potential to detect different manifestations of student disengagement in online learning scenarios. This offers possibilities for attention-aware learning technologies that yield targeted interventions and adapt learning content~\cite[][]{hutt2021breaking, mills2021eye}. Although previous works addressed automatic detection of attention lapses~\cite[][]{bosch2019automatic, lee2022predicting, anonim2023an, singh2023have, gupta2016daisee, kamath2016crowdsourced, gupta2016daisee, kamath2016crowdsourced}, one area for improvement identified by a recent review of these approaches is sample diversity~\cite[][]{kuvar2023detecting}, and one issue in detection is the presence of glasses~\cite[][]{kasneci2024introduction}. As an input to these algorithms, a range of modalities, like eye tracking~\cite[e.g.,][]{kim2018detecting, hutt2017gaze}, have been employed for automated detection. The most scalable and accessible modality is video, as consumer-grade webcams, implemented in most computers, can be employed~\cite[][]{zhao2017scalable, stewart2017face}. Previous research on mind wandering detection and engagement estimation~\cite[][]{bosch2019automatic, lee2022predicting, anonim2023an, buhler2025mapping, singh2023have, gupta2016daisee, kamath2016crowdsourced, gupta2016daisee, kamath2016crowdsourced} commonly relies on image processing and computer vision techniques and train models with video clips captured by webcams and selfie cameras as input samples. Since machine learning requires large amounts of data, student video recordings must first be collected to train a valid model in a centralized manner. However, as images and videos may contain sensitive personal information, storing such data on a central server poses significant privacy concerns, particularly in educational settings~\cite[][]{shokri2015privacy, khosravi2022explainable, nguyen2023ethical}. Facial data collected this way could serve the purpose of student identification, which has already awakened some serious discussions in the educational domain, which would allow automated sorting and classification for students~\cite[][]{andrejevic2020facial}. Previous literature also mentions danger and harm that can be associated with it~\cite[][]{andrejevic2020facial, stark2019facial, crawford2019regulate}. Therefore, in our work, we have a huge emphasis on keeping facial information of the student safe and locally on their devices.

As a collaborative distributed training paradigm, federated learning~\cite[][]{mcmahan2017communication} provides a valid counteractive measure to data sharing risk. In contrast to traditional centralized learning that requires data aggregation, federated learning operates through training localized models on each client using solely its local data. Instead of gathering data from clients, federated learning aggregates locally trained models into a better performing global model and broadcasts the aggregated model across clients. Since data sharing is not needed, federated learning offers a foundational guarantee of user privacy by its nature. A further benefit of federated learning lies in the utilization of local computing power, which means the computation load is distributed among participating entities, and therefore, the burden on the server side can be tremendously reduced.

In this work, we propose a federated learning-based framework for detecting mind wandering, disengagement, and boredom using a user-end camera during remote education and fit a bidirectional long short-term memory (bi-LSTM, ~\cite[]{hochreiter1997long, baldi1999exploiting, schuster1997bidirectional}) network in the framework, which models mind wandering, disengagement, and boredom detection as different video-based binary classification tasks. The separate recognition of these three phenomena is equally crucial in an online learning setting. As online lectures and distance education offer students the flexibility to choose their preferred time, location, and environmental conditions, diverse camera and illumination settings may arise, which presents a notable challenge to machine learning algorithms. A further challenge is the presence of glasses, which can cause reflections and obscure the eyes, preventing the utilization of valuable gaze-related features. To overcome these challenges, we utilize facial and eye gaze features and employ two open-source deep learning models as feature extractors. We also demonstrate a common data quality problem when users wear glasses and suggest a solution. To the end of federated learning, as previous decentralized learning works~\cite[][]{guo2020pedagogical} mainly focus on the application for large-scale data silos like educational institutes to predict students' performance, the novelty of our work lies in the training and inference on the end-user level, which means that each participant with its own device like tablet or computer forms a federated client. On the other hand, while prior research on mind wandering, disengagement, and boredom detection~\cite[][]{bosch2019automatic, lee2022predicting, singh2023have, gupta2016daisee, kamath2016crowdsourced} investigated centralized machine learning methods, none of them considered the privacy of the individuals. Our work utilizes cross-device federated learning and is validated in a user-independent way to better model a real-life use case. We publish our implementation\footnote{\url{https://gitlab.lrz.de/hctl/digital-self-control}\label{url:code}.} for reproducibility. The main contributions of our work can be summarized as follows:
\begin{itemize}
    \item We address three data challenges when developing remote learner state detection algorithms, including disturbance due to glasses, data quality, and data heterogeneity.
    \item We propose a data preprocessing pipeline for video input and further address disturbances caused by glasses by integrating glass-related features. Our algorithm ensures an optimized approach for a broad range of users by delivering comparable model performance for both groups of learners, regardless of whether they wear glasses. 
    \item We utilize cross-device federated learning to approach privacy-preserving distraction detection. By employing advanced federated learning algorithms, we address the problem of data heterogeneity to a large extent. According to our best knowledge, our work is the first to develop federated learning for mind wandering, disengagement, and boredom detection in an educational setting.
    \item We conduct extensive evaluations on five datasets, including three for mind wandering detection, one for engagement analysis, and one for boredom detection. Our benchmarks cover both centralized learning and six federated learning algorithms. Our results depict the great potential of federated learning in replacing centralized learning for privacy preservation.
\end{itemize}

\section{Related Work and Background}\label{chap:relwork} 
\subsection{Remote Learner State Detection}\label{sec:mw_and_engage}
Within the educational domain, the notion of self-regulation~\cite[][]{boekaerts1996self, efklides2011interactions} takes root when students actively and deliberately guide their endeavors and progress towards predetermined educational aims~\cite[][]{bol2011challenges}. Previous works~\cite[][]{azevedo2008externally, dabbagh2004supporting, bol2011challenges, bylieva2021self} underscore that in the context of distance education, the absence of a physical instructor and their assistance tends to lead to a heightened level of mind wandering, disengagement, and boredom from students. However, maintaining sustained focus while studying at home throughout the day presents a formidable challenge for numerous learners~\cite[][]{artino2009beyond, bol2011challenges}. An example of a lack of self-regulation during online lectures is the occurrence of mind wandering, disengagement, or boredom.

Upon initial observation, it is evident that all three mentioned mental states have a consistent negative impact on learning outcomes, potentially resulting in a deficiency of knowledge or even dropout from courses~\cite[][]{cothran2000building, vogel2012definition}. Nevertheless, while mind wandering, disengagement, and boredom phenomena exhibit similarities, they are not identical. Detecting mind wandering presents a more intricate challenge than identifying engagement or boredom. This complexity stems from the fact that mind wandering is a hidden cognitive state; thus, its detection necessitates user input, while in many instances, no discernible behavioral pattern unequivocally indicates its occurrence. Therefore, identifying mind wandering based on facial expressions and behavioral patterns poses a greater challenge compared to recognizing disengagement or boredom.

Since the shift of student attention can occur unintentionally~\cite[][]{schooler2011meta}, developing an algorithm capable of identifying these occurrences from facial expressions and body movements could enhance the learning process's efficacy. 
By focusing on the connection between attention and ocular movement, such as eye-tracking or webcam-based eye-tracking, previous research~\cite[][]{lee2021eyes, rolfs2009microsaccades, gwizdka2019exploring, bergdahl2020engagement, rawat2024computer} underscores the integral relationship between the two: when individuals are engrossed in various activities, their gaze fixations naturally manifest~\cite[][]{lee2021eyes, byrne2023leveraging}. This indicates the significance of monitoring eye movements during reading or video consumption tasks as a pivotal element in remote learner state detection. Although gaze data collected by eye trackers can lead to accurate engagement detection, the presence of eyeglasses or participants having eye disorders could challenge the calibration process and the gaze estimation data, which led to excluding such individuals from the study \cite[][]{raina2016using, dewan2019engagement} or to limited data to evaluate the eye tracking quality~\cite[][]{funke2016eye}. In a user study~\cite[][]{dahlberg2010eye}, the authors experienced a 20\% increase in eye tracking accuracy errors for people wearing glasses. For this subgroup, the eye tracking quality may decrease based on glasses-related factors, such as lens thickness and increased glare~\cite[][]{poole2006eye, funke2016eye}. To accurately detect the gaze of drivers wearing eyeglasses, the authors~\cite[][]{rangesh2020driver} developed a generative adversarial network to remove the eyeglasses from the drivers while preserving the gaze. Although the generated images are promising, the real-time application of this network can be problematic due to its additional computational overhead that would add to our feature extraction and model deployment pipeline.

Continuing the exploration of facial landmark detection for predicting attention lapses, it is pivotal to transition towards discussing datasets and methods employed in engagement and boredom prediction. Previous studies focus on the detection of discerning levels of engagement within educational contexts. This exploration frequently categorizes engagement (and boredom) into four distinct intensities: very low, low, high, and very high ~\cite[][]{whitehill2014faces}. The relevant works rely on facial features and head poses to determine the level of engagement~\cite[][]{gupta2016daisee, kamath2016crowdsourced, singh2023have, kaur2018prediction} and boredom~\cite[][]{gupta2016daisee, kamath2016crowdsourced}, while it can also happen that the authors only labeled the videos as disengagement when a participant was looking away from their monitor and notes~\cite[][]{delgado2021student}. Here, we utilize two behavioral datasets, EngageNet~\cite[][]{singh2023have} and DAiSEE~\cite[][]{gupta2016daisee, kamath2016crowdsourced}, comprising video recordings that observe students' behavior in online course sessions, serving as the training data for our models.

In the EngageNet~\cite[][]{singh2023have} dataset, the authors use the same categories to label engagement and tested different types of network architectures, such as Long Short-Term Memory (LSTM). They also experimented with the input features extracted by the OpenFace~\cite[][]{baltruvsaitis2016openface} framework. The authors achieved the highest model performance when they combined the gaze, head pose, and facial action unit features. However, the size of the dataset is larger compared to mind wandering datasets~\cite[][]{bosch2019automatic, lee2022predicting}, the authors did not remove appearing objects, such as beverages or hands covering the eye region, which can affect the model prediction performance negatively. Although they trained their models on features instead of video data, no measures were taken to ensure the privacy protection of the participants.

Another well-known work, DAiSEE~\cite[][]{gupta2016daisee, kamath2016crowdsourced}, provides labels with the aforementioned categorization corresponding to the boredom, engagement, confusion, and frustration levels of the participants. The authors trained different types of convolutional neural networks for classification and achieved the highest accuracy with LSTMs applied to the 10-second long video sequences and a pre-trained InceptionNet V3 model~\cite[][]{szegedy2016rethinking} applied to the frame level. 

As aforementioned, the paucity of video data exclusively targeting facial expressions during mind wandering has led to a limited number of prior investigations into detecting this complex cognitive phenomenon. In  recent study~\cite[][]{buhler2024detecting}, a new multimodal dataset was introduced that combines eye tracking, facial videos, and physiological signals to improve the detection of both aware and unaware mind wandering during video lecture viewing. With data from 77 university students recorded in Germany, the study demonstrated that video-based facial features, especially when integrated with other modalities, play a key role in identifying mind wandering episodes. The authors achieved an $F_1$ score of 49.3\% (9.56\% above chance) using a multi-layer perceptron (MLP) model applied to OpenFace~\cite[][]{baltruvsaitis2016openface} features extracted from video data, focusing on the combined mind wandering samples. This dataset represents a significant advancement in the field, being the first work focusing on automated detection of aware and unaware mind wandering based on a unique combination of different modalities. A major limitation of the dataset is the absence of participant privacy safeguards in their setup, where all raw data is transmitted and stored centrally without local processing.

A study by \cite{bosch2019automatic}, collected and analyzed video recordings of students deeply engaged in reading scientific texts. The authors used the self-caught method for detecting the mind wandering states and saved them as 10-second-long video samples. They achieved the highest $F_1$ score with support vector machines trained on local binary pattern texture features (15.8\% above chance level), majority vote on all extracted features (15.4\% above chance level). Apart from the limitations associated with the dataset (e.g., the challenge of labeling mind wandering), a more pertinent concern is the oversight of the imperative to preserve students' privacy~\cite[][]{bosch2019automatic}.

\cite{lee2022predicting} introduced an ``in-the-wild'' dataset, capturing participants in Korea as they watched online lecture videos within their preferred settings, such as in their rooms. The authors used a probe-caught method to record mind wandering occurrences, where the participants had to report their attentional state (\textit{Focused} or \textit{Not-Focused}) at pre-defined time points. They achieved the highest $AUROC$ score with 20-second-long videos before the reports using XGBoost. 

A more recent study \cite[][]{anonim2023an} employed those two datasets ~\cite[][]{bosch2019automatic, lee2022predicting} to assess the generalizability of video-based mind wandering detection across tasks and settings, showing the generalization potential for employing transfer-learned latent facial expression features and LSTM models, achieving satisfactory performance for within-dataset (25\% above chance level) and cross-dataset (14\% above chance level) predictions. However, while the authors outperformed or achieved similar results on the datasets as the state-of-the-art work, they did not consider the influence of their model on different learner groups. Furthermore, despite the fact that their model exclusively operated on extracted features rather than exact video frames, the collection of data to a central server still poses a potential threat to the privacy of the participants.

Since the above-mentioned previous works focus on developing an accurate model for mind state detection without preserving participant privacy—often containing sufficient information for participant identification—they do not assess whether their approaches are optimized for a broad range of users or ensure fairness for both groups of learners, regardless of whether they wear glasses. We aim to solve these open questions in our work. Addressing this research gap, we implement federated learning frameworks, which offer a privacy-by-design approach that inherently protects the privacy of individual users.

\subsection{Federated Learning and Applications}
Federated learning is a decentralized collaborative machine learning paradigm that was first proposed in~\cite{mcmahan2017communication}. Compared to conventional centralized learning, where data from clients are gathered to train a central model, in federated learning each client trains its own local model using its local data from the same starting point, and a server periodically aggregates the locally trained models. Since no data exchange is required, federated learning has great potential to reduce data risk regarding privacy and security. In the rest of this work, we focus on classification tasks using embedding-based networks. In a multi-class classification task with $K \in \mathbb{N}^+$ classes and $N \in \mathbb{N}^+$ clients, define the local dataset of each client as $D_1, D_2, ..., D_i, ..., D_N$, with $i \in [N]$. A centralized learning approach commonly starts with collecting client local datasets and forming a global dataset $D_G$, namely: 
\begin{equation}
D_G = \biguplus_{n=1}^{N} D_n \text{, with } |D_G| = \sum_{n=1}^{N} |D_n|     
\end{equation}
The goal of centralized learning is often to look for an optimal model $\theta^C$ that minimizes the empirical risk on the global dataset, i.e.: 
\begin{equation}
\theta^C = \operatornamewithlimits{argmin}_\theta \mathbb{E}_{(x,y) \sim D_G} [\ell(x, y, \theta)],
\end{equation}
where $\ell(x, y, \theta)$ describes the objective function evaluated on the sample $(x, y)$ using model parameters $\theta$. In contrast to centralized learning, federated learning searches for a model $\theta^{FL}$ by solving the objective collaboratively among the clients:
\begin{equation}
\theta^{FL} = \operatornamewithlimits{argmin}_\theta \sum_{i=1}^{N} \frac{|D_i|}{|D_G|} \mathbb{E}_{(x,y) \sim D_i} [\ell(x, y, \theta)]
\end{equation}
Typically, the workflow of federated learning consists of three main steps. First, a server initializes a global model and forwards this model to each client. Then, the clients train the received models using their local datasets and resources for several epochs and send the trained models back to the central server. Afterward, the server aggregates the collected trained models and sends the latest global model to clients.

Along with the concept of federated learning, the very first aggregation algorithm, FedAvg~\cite[][]{mcmahan2017communication}, was introduced. On the client side, FedAvg trains models with stochastic gradient descent (SGD), while on the server side, it applies (weighted) average over client models. Many follow-up works have been contributed to federated learning to improve model performance, robustness, privacy guarantees, etc. For instance, FedOpt~\cite[][]{reddi2020adaptive} incorporates momentum-based adaptive optimization into federated aggregation. By handling local model updates as pseudo gradients, FedOpt allows the use of advanced optimizers like Adagrad~\cite[][]{lydia2019adagrad}, Adam~\cite[][]{kingma2014adam}, and Yogi~\cite[][]{zaheer2018adaptive} on the server side instead of naive weighted average. One of the biggest challenges in federated learning is that the data are often not independently identically distributed (non-iid) among clients, and many efforts have been devoted to solving the non-iid problem, such as FedProx~\cite[][]{li2020federated}, SCAFFOLD~\cite[][]{karimireddy2020scaffold}, FedMA~\cite[][]{wang2020federated}, FedAvgM~\cite[][]{hsu2019measuring}, MOON~\cite[][]{li2021model}, and FedProc~\cite[][]{mu2023fedproc}. Two special works in this field are FedAwS~\cite[][]{yu2020federated} and FedDC~\cite[][]{kamp2021federated}, in which the authors tried to address extreme data heterogeneity. The former deals with the scenario that each client had merely data from one class, while the latter works even when some clients have only two samples. More recently, a novel method called TurboSVM-FL~\cite[][]{wang2024turbosvmfl} was proposed, which targets both speeding up convergence and tackling client drift, without posing any additional computation burden on the client side. The application of federated learning to approach ubiquitous computing has also experienced a giant stride in recent years. For instance, FDAS~\cite[][]{gong24privacy} focuses on concurrent alignment at both domain and semantic levels to improve the semantic quality of the aligned features, and EchoPFL~\cite[][]{Li24echopfl} uses a coordination mechanism for asynchronous personalized federated learning while  FedCHAR~\cite[][]{Li24hierarch} is a personalized federated learning framework with a hierarchical clustering method. For an extended evaluation of federated learning methods, we refer the reader to the surveys~\cite[][]{tan2022towards, wang2021field}. In summary, federated learning enhances privacy by training models on users' devices, keeping sensitive data local. It also supports personalization by adapting to individual user characteristics and environmental factors, ensuring more relevant results. Moreover, it reduces data transfer costs, maintains data security, and allows for scalable deployment across diverse user groups without the need for centralized data collection.

Concerns about data privacy weigh heavily for educational institutions, prompting meticulous consideration regarding data aggregation practices based on the nature of raw data. In online learning, especially when accessing student video data~\cite[][]{pattanasethanon2012human}, we place a strong emphasis on privacy. The pervasive potential of remote learner state detection algorithms in education amplifies the urgency of shielding user-sensitive data, such as facial video content, from centralized exposure. Consequently, the strategic employment of federated learning emerges as a means to meticulously preserve privacy throughout model training and evaluation phases~\cite[][]{wang2025cyclesl}. However, adopting federated learning in education technologies focusing on possible real-time support remains unexplored. Notable among the limited endeavors is the work~\cite[][]{guo2020pedagogical}, wherein the authors introduced the Federated Education Data Analysis framework. This study harnessed the data from diverse institutes, drawn from students, to predict academic performance and potential dropouts. Their approach designates each institute as a client, aligning with the cross-silo federated learning domain, which is essentially different from the settings in distance education where each client represents an individual end user, not an institution, placing our work squarely in the cross-device federated learning category. \citet[][]{van2024federated} compared three machine learning architectures—local, federated, and central learning—across different educational prediction tasks and found that federated learning consistently performed on par with central learning, while local learning remained competitive up to 20 clients. \citet[][]{fachola2023federated} highlighted the potential of federated learning in learning analytics, particularly for student dropout prediction, demonstrating that it achieves comparable performance to centralized models while addressing legal and ethical concerns related to sensitive data processing. While previous applications of federated learning in education have focused primarily on educational analytics, such as performance prediction and dropout detection, its potential for real-time learning support during lecture watching remains largely unexplored. In this context, leveraging federated learning for mind wandering, boredom, and disengagement detection presents a promising approach because it enables model training directly on students' devices without requiring raw video data to be shared, thus preserving privacy. Unlike transfer learning~\cite[][]{gardner2023cross, weiss2016survey}, which typically relies on a pre-trained model that may not generalize well across diverse learning environments~\cite[][]{chen2022rethinking}, federated learning allows models to adapt to varying conditions such as different lighting, camera angles, and engagement patterns, ensuring both personalization and user data confidentiality.

\section{Challenges}

\subsection{Disturbance of Glasses} \label{chal_glass}
\begin{figure}[t]
	\centering
	\subfloat[User without glasses.]{\includegraphics[height=3cm, keepaspectratio]{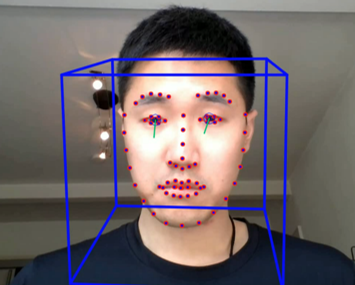}
    \label{fig:noglass}}
    \hspace{5mm}
	\subfloat[User with glasses.]{\includegraphics[height=3cm, keepaspectratio]{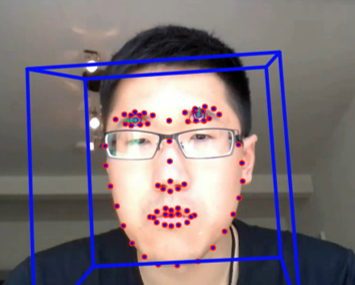}
	\label{fig:glass}}
	\hfil
	\caption{Example of correct and incorrect eyes detection with Openface~\cite[][]{baltrusaitis2018openface}. In both cases, OpenFace yields a confidence value over 0.97.}
    \label{fig:glasses_OF}
\end{figure}

Students engaged in online learning typically watch pre-recorded lectures or live-streamed classes on their monitors. Since a large portion of the population requires vision correction, and most of them rely on eyeglasses, this factor is worth considering. In the U.S., 75.6\% of adults use some form of vision correction, with 63.7\% of them wearing eyeglasses~\cite[][]{tvc2021org}. Similarly, in Europe, 48\% of the population used eyeglasses in 2020~\cite[][]{european2020share}. Given these figures, it is important to examine whether wearing eyeglasses affects the detection of remote learner states. Although highly accurate eye trackers are used for gaze estimation, the presence of eyeglasses makes the gaze estimation task challenging~\cite[][]{poole2006eye, funke2016eye}. Hence, it is crucial to acknowledge that the presence of eyeglasses could further complicate the gaze estimation task when utilizing a webcam. Also, by collecting video data during online lectures, the screen reflection on participants' glasses might make their eyes unrecognizable and untraceable in a selected frame sequence, while the confidence provided by the OpenFace~\cite[][]{baltrusaitis2018openface} framework still stays above 0.97 (Figure~\ref{fig:glasses_OF}).

\subsection{Data Quality}
To effectively assist as many participants enrolled in an online course as possible with their behavioral and cognitive engagement, it is imperative that we leverage the resources available to them. Given that most students rely on their computers or laptops for their learning endeavors~\cite[][]{miklyaeva2020self}, the most accessible and practical tool at their disposal is the built-in or USB webcam. However, the inherent limitations of webcam resolution introduce a significant obstacle to accurate facial landmark detection.

Furthermore, the flexibility of online learning enables students to partake in courses from a range of environments, such as their personal rooms, outdoor spaces, etc.~\cite[][]{miklyaeva2020self}. This wide range of settings results in variations in lighting conditions and backgrounds, even for the same individuals, posing an additional complexity. These diverse settings not only challenge the efficacy of remote learner state detection but can also impede the overall effectiveness of the students' learning process itself~\cite[][]{pattanasethanon2012human}. Moreover, the positioning of the camera, compounded by students' natural tendency to move around and engage in actions like standing, stretching, or taking sips of beverages, often leads to transient moments when their faces become temporarily obscured. These transient instances greatly compound the intricacies of any remote learner state detection endeavor.

In summary, the intricate interplay of environmental variables, device limitations, user behavior, and labeling of datasets presents a formidable challenge to the seamless execution of remote learner state detection in online learning. To address this issue, we compare different federated learning frameworks to additionally also preserve participants' privacy.

\subsection{Data Heterogeneity}
\begin{figure}[t]
	\centering
	\subfloat[Histogram of numbers of samples per user in the  Colorado, Korea, Germany, EngageNet, and DAiSEE datasets.]{\includegraphics[height=4.1cm, keepaspectratio]{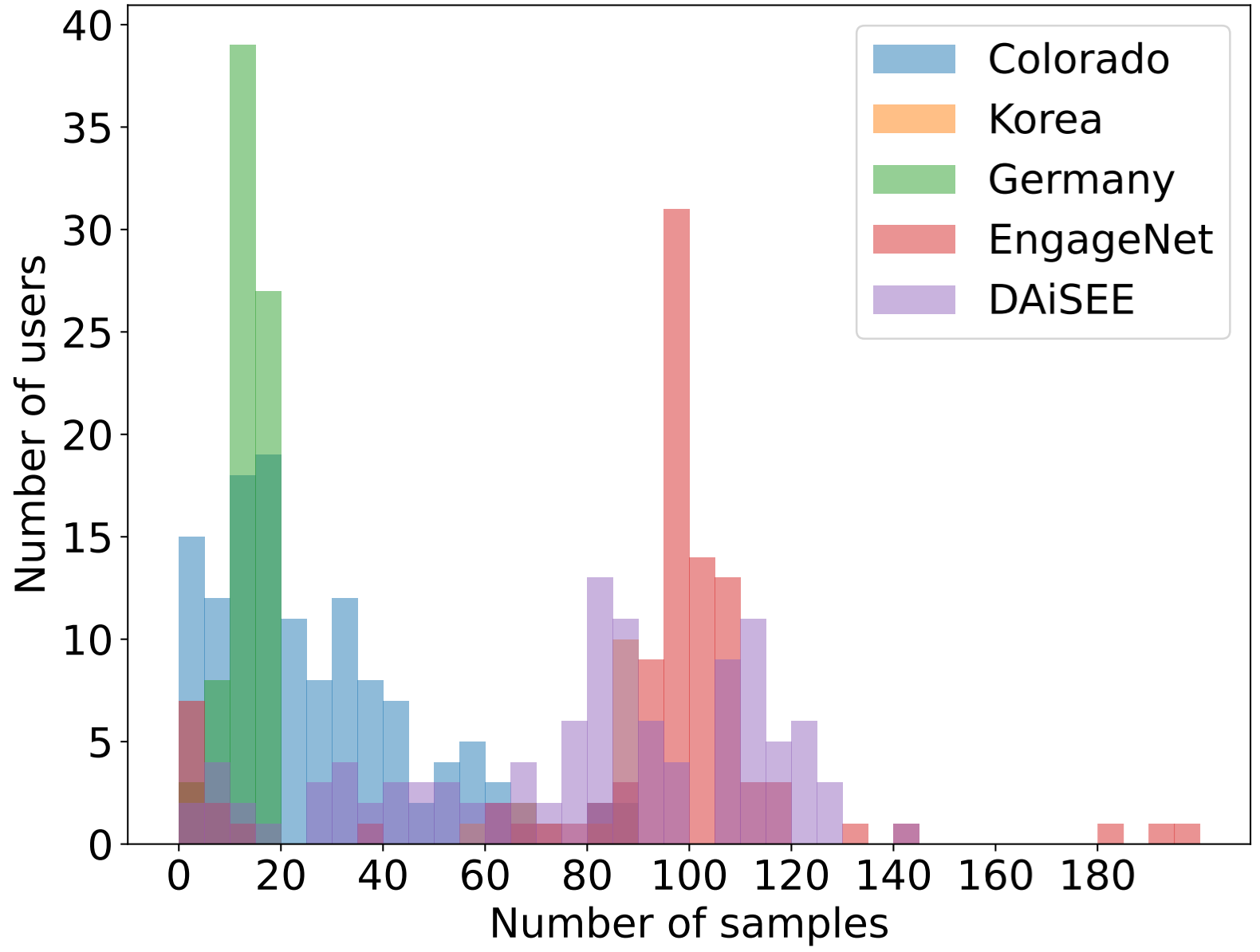}
    \label{fig:histo1}}
    \hspace{15mm}
	\subfloat[t-SNE~\cite{van2008visualizing} representations of the Korea glass features. Warm colors: people with glasses; cold colors: people without glasses.]{\includegraphics[height=3.9cm, keepaspectratio]{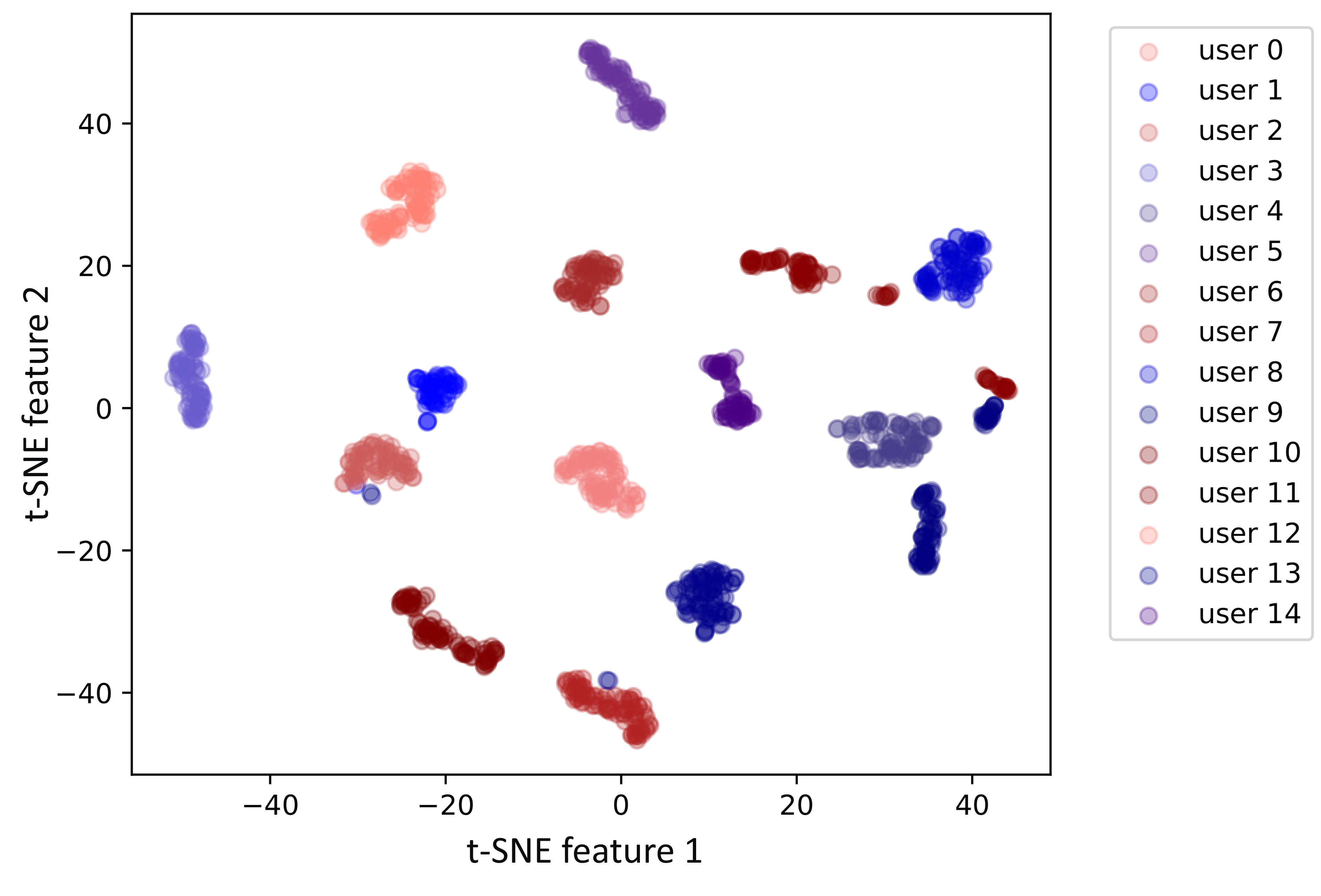}
	\label{fig:tsne}}
	\hfil
	\caption{Dataset visualization based on the number of samples per client (a), and hidden representations (b)}
    \label{fig:dataset_sum}
\end{figure}

Data heterogeneity across client local datasets is a common challenge in federated learning. It can lead to a phenomenon called client drift~\cite[][]{karimireddy2020scaffold} where the client models are trapped in local optima instead of global optima, which can significantly influence the performance of the aggregated global model. In many existing federated learning works, non-iid distribution among clients is often manually simulated on well-studied datasets, such as MNIST~\cite[][]{minst2012li} and CIFAR~\cite[][]{krizhevsky2009learning}. In practical applications, data heterogeneity can be however exacerbated and much more complicated. For instance, Figure~\ref{fig:histo1} depicts the data distribution in the Colorado~\cite[][]{bosch2019automatic}, Korea~\cite[][]{lee2022predicting}, Germany~\cite[][]{buhler2024detecting}, EngageNet~\cite[][]{singh2023have}, and DAiSEE~\cite[][]{gupta2016daisee, kamath2016crowdsourced},  datasets. In the Colorado~\cite[][]{bosch2019automatic} dataset, while an average amount of samples per user of around 25 can be observed, 15 out of 130 users have merely 4 or fewer samples. In comparison to this, 4 participants have over 80 samples. Although the overall positive sample rate among 130 users reaches 30\%, 8 of them have only mind wandering samples, while 15 participants never reported mind wandering during experiments. A further example is given in Figure~\ref{fig:tsne}, which is a 2D t-SNE~\cite[][]{van2008visualizing} visualization of feature vectors of the Korea dataset extracted by a glass detection network (ResNet18~\cite[][]{he2016deep}, fine-tuned on the MeGlass~\cite[][]{guo2018face} dataset). Although this dataset contains only 15 users and all users have similar sample sizes, they still form 15 distinct clusters in the latent space. The presence of glasses cannot be determined by analyzing the locations of feature points for each participant, as the visualized features of samples from participants with and without glasses do not form distinct separable patterns.

\section{Methodology}\label{chap:methodology}

\begin{figure*}[t]
	\centering
	\includegraphics[width=\textwidth, keepaspectratio]{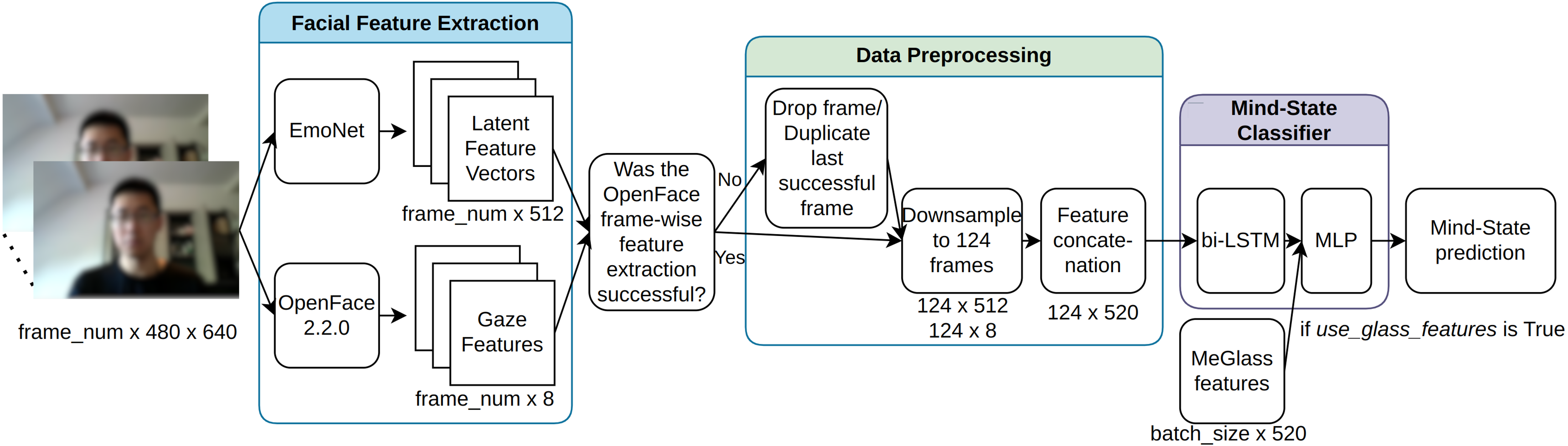}
	\caption{Model architecture. After extracting the EmoNet~\cite[][]{toisoul2021estimation} and OpenFace~\cite[][]{baltrusaitis2018openface} features, we analyze each frame, whether OpenFace feature extraction was successful (this is saved for each frame and shows OpenFace's confidence in its prediction). Next, we downsample the frames with detected faces on them and create the input vector for the neural network, which consists of two main parts: bi-LSTM and MLP layers.}
    \label{fig:model}
\end{figure*}

To detect remote learner states, we develop a neural network focusing on a binary classification problem. Since many available disengagement detection datasets~\cite[][]{bosch2019automatic, lee2022predicting, singh2023have, gupta2016daisee, kamath2016crowdsourced} contain video data, we utilize recurrent neural networks as the backbone model, as suggested in~\citet[][]{singh2023have}. Particularly, we adopt the model structure as in~\cite[][]{anonim2023an} and employ a 3-layer bidirectional long short-term memory model (bi-LSTM,~\cite[][]{hochreiter1997long, baldi1999exploiting, schuster1997bidirectional, anonim2023an}) with 100 hidden units to capture time dependencies. We also implemented a bagging ensemble~\cite[][]{breiman1996bagging} with the bi-LSTM model to explore the benefits of combining multiple models for enhanced detection performance. We applied user-wise sampling to establish a benchmark that is more directly comparable to the federated learning algorithms. To improve the representation ability of the class embeddings, we follow the strategy suggested in~\cite[][]{chen2020simple, anonim2023an} and use a single-layer linear projection (Figure~\ref{fig:model}) to map the output of bi-LSTM to class probabilities. As for input, we employ two open-source networks to extract features from video frames, namely EmoNet~\cite[][]{toisoul2021estimation} and OpenFace~\cite[][]{baltrusaitis2018openface} (latest version 2.2.0), which are publicly available. The former is designed for emotion classification, while the latter is trained for the detection of facial landmarks, coordinates, gaze, and head pose. The overall pipeline is rather lightweight and can be run on end devices that are not equipped with dedicated graphic processors. 

To enhance model performance for participants with glasses and improve accuracy for both those with and without glasses, we concatenate a feature vector from an eyeglass detection network with the output of the bi-LSTM to adjust the significance of these features. This way, we let the network decide whether the reflection in the glasses makes the gaze feature extraction irrelevant and whether the model should rely on the other features. These features could be optionally added before the linear projection layer (Figure~\ref{fig:model}).

To address the privacy concerns regarding the facial data collection of students, we integrate the aforementioned model into a federated learning strategy, namely, handling the participants as separate clients. We evaluate the model performance both in centralized and decentralized learning. For the latter, we benchmark six different federated learning algorithms, including FedAvg~\cite[][]{mcmahan2017communication}, FedAdam~\cite[][]{reddi2020adaptive}, FedProx~\cite[][]{li2020federated}, MOON~\cite[][]{li2021model}, FedAwS~\cite[][]{yu2020federated}, and TurboSVM-FL~\cite[][]{wang2024turbosvmfl}. Among these six algorithms, FedAvg is the vanilla federated learning aggregation method, while FedAdam belongs to a family called FedOpt~\cite[][]{reddi2020adaptive} that aims to benefit federated learning with adaptivity. FedProx, MOON, FedAwS, and TurboSVM-FL attempt to address data heterogeneity. The former two bring additional penalty-based loss terms to the local training on the client side, while the latter two improve model aggregation on the server side. It should be noticed that although, in general, the output layer for a binary classification problem commonly has only one logit, we use two logits as the class embeddings of both classes are needed for some federated learning algorithms, such as FedAwS and TurboSVM-FL. Subsequently, the activation function for the logit layer is not sigmoid but rather softmax.

\section{Experiments}\label{chap:experiments}
For each centralized and decentralized algorithm, we repeated the experiment five times with a different random seed in the range of $\{0, 1, 2, 3, 4\}$ each time, and subsequently, we reported the mean and standard deviation for each metric over five seeds after individual fine-tuning and optimization. We report our main findings using the binary $F_1$ score, above chance level of $F_1$ score (its sign shows whether it is above (positive) or below (negative) chance level), precision and recall (weighted $F_1$ score, accuracy, and area under the curve (AUC) is reported in the Appendix~\autoref{tab:FL_SUM_additional}), as these metrics are the most commonly reported in previous works~\cite[][]{kuvar2023detecting}. We applied a fixed, randomized 90\%-10\% train-test split, in a user-independent way, meaning that we had a hold-out set of users with all their data spared for test and validation instead of hold-out samples per user~\cite[][]{wang2021field}. More details about our experiments, such as environment, data distribution, and hyperparameter tuning, can be found in the Appendix~\ref{chap:hparam}. We published our implementation and instructions for reproducing our experiments\textsuperscript{\ref{url:code}}.

\subsection{Datasets}
We evaluated our algorithm separately on five recent datasets, including Colorado~\cite[][]{bosch2019automatic}, Korea~\cite[][]{lee2022predicting}, Germany~\cite[][]{buhler2024detecting}, EngageNet~\cite[][]{singh2023have}, and DAiSEE~\cite[][]{gupta2016daisee, kamath2016crowdsourced} (asummary is presented in Table~\ref{tab:datasets}). All five datasets are either publicly available, were provided by the authors on demand. All datasets were raised from remote learning, where students sat in front of their screens while their background and illumination could vary on a large scale. The first three datasets are labeled for mind wandering detection, the fourth one is for engagement detection, and the last dataset provides labels for the prediction of the level of engagement, boredom, confusion, and frustration. Especially, although the DAiSEE dataset contains labels for four different mind states, we only focus on the binary classification of boredom, since the label distributions of other categories are highly unbalanced. In the Colorado, Korea, and Germany datasets, video labels are binary, i.e., labeling mind wandering and non-mind wandering. In the Germany dataset, mind wandering labels are distinguished between aware and unaware episodes; however, consistent with the approach taken by the original authors, these categories can be merged to facilitate a binary classification task. In contrast, the EngageNet and DAiSEE datasets provide labels across four gradations: very low, low, high, and very high. We consolidated these four gradations into a binary classification framework for our analysis, namely grouping very low and low, and high and very high labels together. The binary simplification of the engagement and boredom levels in these datasets is feasible and also suggested~\cite[][]{gupta2016daisee}. All videos were standardized to a resolution of $640\times 480$, trimmed to a duration of ten seconds, and adjusted to comprise 124 frames prior to their introduction to the network. Participants with four or fewer samples were consequently excluded from further analysis. After data preprocessing, we excluded 6 participants from the EngageNet and 12 participants from the Colorado datasets. 
\begin{table*}[t] 
    \centering
    \caption{Dataset statistics (Colorado~\cite[][]{bosch2019automatic}, Korea~\cite[][]{lee2022predicting}, Germany~\cite[][]{buhler2024detecting}, EngageNet~\cite[][]{singh2023have}, and DAiSEE~\cite[][]{gupta2016daisee, kamath2016crowdsourced}). The \textit{positives} column refers to the number of disengagement, mind wandering, and boredom samples in the corresponding datasets.}
    \begin{tabular}{p{16mm} p{10mm} p{8mm} p{8mm} p{8mm} p{10mm} p{16mm} p{5mm} p{8mm}}
    \toprule
        Dataset & Task & Sam-ples & Posi-tive & Users & Glasses & Report Type & FPS & Env. \\
        \midrule
        Colorado & reading & 3303 & 995 & 130 & 12 & self-caught  & 12.5 & lab \\
        Korea & watching & 1220 & 206 & 15 & 7 & probe-caught  & 30 & home \\
        Germany & watching & 939 & 256 & 77 & 24 & probe-caught  & 30 & lab \\
        EngageNet & watching & 9004 & 2803 & 99 & 33 & question-naire  & 30 & diverse \\
        DAiSEE & watching & 8925 & 2253 & 112 & 54 & annotator & 30 & diverse \\
        \bottomrule
    \end{tabular}
    
    \label{tab:datasets}
\end{table*}

Although these five datasets exhibit similarities, they are annotated for the recognition of specific mental states. Our approach involves applying the same model structure to all five datasets. Consequently, individually trained models are exclusively equipped for the binary classification of the phenomena labeled in each dataset. Our objective is not to create a framework capable of simultaneously detecting disengagement, boredom, and mind wandering in a multi-label scenario. Instead, we aim to devise a framework capable of identifying these manifestations of disengagement individually. We extracted features by EmoNet~\cite[][]{toisoul2021estimation} and Openface~\cite[][]{baltrusaitis2018openface} and combined these features as our input. A brief discussion about the necessity of these features and the exact feature extraction is given in the Appendix~\ref{app:facial_features}. 

\subsection{Focusing on People with Glasses}
The main challenge for people with glasses is the possible occurrence of reflection on the spectacles, which could hide the eyes. Thus, gaze retrieval becomes difficult. We focus on developing an algorithm in centralized learning that can effectively detect disengagement for students with or without glasses, ensuring it is optimized for a broad range of learners.

\begin{table*}[t]
    \centering
    \scriptsize
    \caption{Centralized learning results with and without MeGlass~\cite[][]{guo2018face} features on the EngageNet~\cite[][]{singh2023have}, Colorado~\cite[][]{bosch2019automatic}, Korea~\cite[][]{lee2022predicting} and DAiSEE~\cite[][]{gupta2016daisee, kamath2016crowdsourced} datasets. \textit{All} rows show overall model performance, while \textit{Glasses} and \textit{No-Glasses} rows refer to the achieved model performance on the users who wear and do not wear glasses separately.}
    \begin{tabular}{p{5mm} p{10mm} p{7mm} p{14mm} p{14mm} p{14mm} p{14mm} p{14mm}}
    \toprule
         &\multicolumn{2}{c}{Performance} & Colorado & Korea & Germany & EngageNet & DAiSEE \\
        \midrule
        \multirow{10}{*}{\begin{sideways} \thead{Baseline \\ bi-LSTM } \end{sideways}} & \multirow{4}{*}{All} & $AC$[\%] & \textbf{22.6} & \textbf{4.7} & 20.9 & 26.6 & 0.5 \\
        & & $F_1$[\%] & \textbf{45.9$\pm$6.7} & 2\textbf{0.8$\pm$11.3} & \textbf{52.0$\pm$4.1} & 49.4$\pm$5.5 & 25.6$\pm$9.9 \\
        & & $Prec$[\%] & 37.7$\pm$4.4 & 22.8$\pm$9.9 & 46.3$\pm$6.3 & 61.4$\pm$4.4 & 16.4$\pm$0.8 \\
        & & $Rec$[\%] & 59.6$\pm$13.3 & 21.1$\pm$13.3 & 60.5$\pm$8.7 & 42.0$\pm$8.1 & 58.6$\pm$2.5 \\
        \cmidrule{2-8}
         & \multirow{3}{*}{Glasses} & $F_1$[\%] & 24.6$\pm$3.0 & \textbf{24.2$\pm$14.3} & \textbf{58.7$\pm$4.9} & 36.7$\pm$15.1 & \textbf{33.5$\pm$1.2} \\
        & & $Prec$[\%] & 15.0$\pm$2.3 & 26.6$\pm$18.2 & 48.9$\pm$7.4 & 59.4$\pm$11.3 & 21.0$\pm$0.7 \\
        & & $Rec$[\%] & 70.9$\pm$5.7 & 27.0$\pm$19.5 & 75.0$\pm$8.3 & 27.3$\pm$13.5 & 82.2$\pm$6.4 \\
        \cmidrule{2-8}
         & \multirow{3}{*}{\thead{No- \\ Glasses}} & $F_1$[\%] & \textbf{53.1$\pm$7.4} & \textbf{13.8$\pm$7.0} & \textbf{48.0$\pm$6.3} & \textbf{57.9$\pm$2.5} & 5.6$\pm$4.7 \\
        & & $Prec$[\%] & 49.9$\pm$2.9 & 36.9$\pm$39.3 & 44.9$\pm$5.9 & 61.7$\pm$8.3 & 3.7$\pm$2.8 \\
        & & $Rec$[\%] & 58.2$\pm$14.4 & 10.8$\pm$4.2 & 53.8$\pm$15.6 & 55.4$\pm$4.4 & 13.0$\pm$14.2 \\
        \toprule
        \multirow{10}{*}{\begin{sideways} \thead{MeGlass \\ features} \end{sideways}} & \multirow{4}{*}{All} &  $AC$[\%] & 21.6 & -2.6 & 12.9 & \textbf{31.9} & \textbf{2.3} \\
        & & $F_1$[\%] & 45.2$\pm$3.9 & 14.7$\pm$10.5 & 47.1$\pm$14.0& \textbf{53.1$\pm$3.5} & \textbf{26.9$\pm$0.6} \\
        & & $Prec$[\%] & 39.0$\pm$3.0 & 11.0$\pm$8.7 & 42.2$\pm$8.0 & 52.3$\pm$7.2 & 16.5$\pm$0.6 \\
        & & $Rec$[\%] & 54.1$\pm$7.0 & 33.3$\pm$37.5 & 60.5$\pm$30.1 & 55.8$\pm$9.9 & 72.3$\pm$7.2 \\
        \cmidrule{2-8}
         & \multirow{2}{*}{Glasses} & $F_1$[\%] & \textbf{26.9$\pm$3.1} & 16.3$\pm$17.4 & 54.3$\pm$7.2 & \textbf{58.7$\pm$19.1} & 31.1$\pm$0.5 \\
        & & $Prec$[\%] & 17.5$\pm$3.1 & 17.3$\pm$16.0 & 46.7$\pm$8.2 & 74.2$\pm$4.6 & 20.1$\pm$0.7 \\
        & & $Rec$[\%] & 63.6$\pm$14.4 & 26.1$\pm$42.0 & 68.3$\pm$16.0 & 51.9$\pm$24.8 & 69.8$\pm$10.4 \\
        \cmidrule{2-8}
         & \multirow{3}{*}{\thead{No- \\ Glasses}} & $F_1$[\%] & 50.1$\pm$4.1 & 12.8$\pm$10.4 & 42.4$\pm$20.5 & 48.8$\pm$13.0 & \textbf{21.6$\pm$2.3} \\
        & & $Prec$[\%] & 48.9$\pm$1.6 & 8.4$\pm$7.9 & 39.8$\pm$12.2 & 42.2$\pm$13.3 & 12.6$\pm$1.3 \\
        & & $Rec$[\%] & 52.9$\pm$8.8 & 46.2$\pm$43.2 & 56.9$\pm$38.0 & 59.4$\pm$13.5 & 77.0$\pm$12.6 \\
        \bottomrule
    \end{tabular}
    
    \label{tab:non_FL,MeGlass}
\end{table*}

Since the presence of eyeglasses could falsify the extracted OpenFace features without being detected as unsuccessful by the algorithm itself, as demonstrated in section~\ref{chal_glass}, the accuracy of the mind-state prediction could decrease. To address this, we give additional input to the network to decide whether the gaze features and eye region contain valuable information regarding disengagement detection, instead of ignoring the gaze features when the participants are wearing glasses. This way, gaze features could also contain meaningful information for people with glasses. As an additional input source, we extracted glass-related features with a pre-trained ResNet18~\cite[][]{he2016deep} model that we fine-tuned on a publicly available glass-involved dataset, MeGlass~\cite[][]{guo2018face}. It achieved a 99.57 \% $F_1$ score on the test set. The extracted glass-related features with this network are of dimension 256. 

As the presence of glasses is constant for each participant and the glass features do not vary much in the ten-second-long videos, we only concatenated the means and standard deviations of glass features over time to the output of the bi-LSTM model. Unfortunately, the ratio between participants with and without glasses was unbalanced (Table~\ref{tab:datasets}). We compared our results (Table~\ref{tab:non_FL,MeGlass}) to a baseline bi-LSTM model, which we introduced in chapter~\ref{chap:methodology}. For a fair comparison, it is crucial to use the same baseline model, as this ensures a realistic evaluation of performance across the settings. Based on our results, we observe that for the non-mind-wandering datasets, the inclusion of MeGlass features led to a slight improvement in model performance. In contrast, for the mind-wandering datasets, the baseline bi-LSTM either outperformed the models enhanced with MeGlass features or achieved comparable results. This suggests that the added features may not consistently contribute to predictive performance across different contexts.

It is important to note that the potential benefit of the MeGlass features should be interpreted with caution, as these features have yet to be thoroughly evaluated on more balanced datasets. For example, in the test sets from Korea, there is only one participant with glasses; in the Colorado set, only three; and in the Germany set, only two. This limited representation could bias the observed effects, as the models may not have sufficient variability to generalize the influence of glasses-related features. Future work should validate these features in larger and more representative samples, where the presence of glasses and other relevant attributes are adequately distributed across the data. This would allow for a clearer understanding of whether MeGlass features provide meaningful additional information beyond what is captured by the baseline models.

\subsection{Results of User-Independent Federated Learning}
We trained and evaluated our model in cross-device federated learning with user-independent splits, assigning each user as a client and performing the train-test split at the client level rather than the sample level. More precisely, we used a hold-out set of clients for the different sets, rather than allocating a proportion of test samples within each client.~\cite[][]{wang2021field}. The primary motivation behind performing user-independent validation is its superior validity in handling unseen data, which makes it a more suitable representation for real-world applications. Additionally, modeling in a user-independent context is notably more demanding compared to using user-dependent data splits. To benchmark our methods, we conduct experiments using a bi-LSTM model and a bagging approach (Appendix~\ref{sec:bagging}). The bagging ensemble combines multiple bi-LSTM models trained on user-wise samples, effectively simulating the cross-device federated learning scenario to create a comparable benchmark for evaluating federated learning performance. The key difference between bagging and federated learning in this case lies in how final predictions are generated: bagging combines the outputs of individual models through majority voting across models trained on separate user data, whereas federated learning aggregates the parameters of these separately trained models using various strategies to form a single, unified model that generates the final predictions.

We simulated the federated learning scenario with virtual clients, where the data was stored on a centralized server. Each client trains local models for eight epochs before aggregation and the learning rates for the clients were tuned for FedAvg~\cite[][]{mcmahan2017communication}. We compared six different federated learning models to centralized learning to systematically evaluate the impact of various aggregation and optimization strategies on model performance. This comparison provides a comprehensive analysis of how different federated learning approaches handle decentralized data distribution and whether they can match or exceed the performance of a traditionally trained centralized model.

\begin{table*}[t]
    \centering
    \scriptsize
    \caption{Federated learning results on the five datasets (Colorado~\cite[][]{bosch2019automatic}, Korea~\cite[][]{lee2022predicting}, Germany~\cite[][]{buhler2024detecting}, EngageNet~\cite[][]{singh2023have}, and DAiSEE~\cite[][]{gupta2016daisee, kamath2016crowdsourced}). All federated learning results are obtained with bi-LSTM model. The $F_1$ score is reported with the ``binary'' setting. \textbf{AC stands for above chance level $F_1$ score difference.}}
    \begin{tabular}{p{16mm} p{9mm} c c c c c}
    \toprule
        \multicolumn{2}{c}{Performance} & Colorado & Korea & Germany & EngageNet & DAiSEE \\
        \midrule
        Chance & - & 30.1 & 16.9 & 39.3 & 31.1 & 25.2 \\
        \midrule
         \multirow{4}{*}{\thead{Centralized \\ Learning \\ (bi-LSTM)}} & $F_1$[\%] & 43.9$\pm$1.4 & 24.2$\pm$6.8 & 50.0$\pm$4.5 & 51.4$\pm$6.0 & 25.7$\pm$0.6 \\
          & $AC$[\%] & 19.7 & \textbf{8.9} & 17.6 & 29.5 & 0.7 \\
         & $Prec$[\%] & 30.4$\pm$2.5 & 15.2$\pm$6.6 & 44.1$\pm$5.3 & 55.7$\pm$2.3 & 16.3$\pm$0.7 \\
          & $Rec$[\%] & 81.1$\pm$11.6 & 76.0$\pm$14.6 & 57.9$\pm$3.7 & 48.6$\pm$10.0 & 61.6$\pm$6.8 \\
         \midrule
         \multirow{4}{*}{\thead{Centralized \\ Learning \\ (bagging)}} & $F_1$[\%] & \textbf{48.1$\pm$1.9} & \textbf{24.8$\pm$10.6} & 41.7$\pm$9.9 & 44.3$\pm$6.5 & 25.3$\pm$1.1 \\
         & $AC$[\%] & 25.8 & 9.5 & 4.0 & 19.2 & 0.1 \\
         & $Prec$[\%] & 33.2$\pm$2.9 & 31.4$\pm$14.7 & 46.0$\pm$3.9 & 67.9$\pm$12.5 & 16.6$\pm$0.8 \\
         & $Rec$[\%] & 88.6$\pm$7.1 & 28.0$\pm$15.9 & 41.1$\pm$16.4 & 36.7$\pm$16.7 & 52.5$\pm$2.7 \\
         \midrule
        \multirow{4}{*}{FedAvg} & $F_1$[\%] & 46.2$\pm$5.4 & 22.1$\pm$2.8 & 52.6$\pm$3.6 & 57.4$\pm$4.6 & 26.2$\pm$1.8 \\
          & $AC$[\%] & \textbf{23.0} & 6.3 & 21.9 & 38.2 & 1.3 \\
          & $Prec$[\%] & 41.1$\pm$5.8 & 13.4$\pm$2.4 & 43.0$\pm$3.4 & 51.9$\pm$3.1 & 16.3$\pm$1.4 \\
          & $Rec$[\%] & 53.9$\pm$9.7 & 69.3$\pm$12.1 & 71.6$\pm$19.8 & 65.3$\pm$10.4 & 67.3$\pm$6.8 \\
        \midrule
        \multirow{4}{*}{FedAdam} & $F_1$[\%] & 41.4$\pm$5.7 & 16.9$\pm$6.2 & 51.4$\pm$4.2 & 49.1$\pm$4.7 & \textbf{26.4$\pm$0.7} \\
          & $AC$[\%] & 16.2 & 0.0 & 19.9 & 26.1 & \textbf{1.6} \\
          & $Prec$[\%] & 40.4$\pm$2.3 & 10.5$\pm$3.4 & 45.1$\pm$2.0 & 57.4$\pm$5.9 & 16.0$\pm$0.9 \\
          & $Rec$[\%] & 44.1$\pm$12.9 & 52.0$\pm$35.7 & 60.5$\pm$10.0 & 43.4$\pm$6.4 & 79.6$\pm$16.9 \\
        \midrule
        \multirow{4}{*}{FedAwS} & $F_1$[\%] & 46.2$\pm$4.4 & 23.0$\pm$9.9 & \textbf{55.7$\pm$3.0} & 55.7$\pm$3.1 & 25.0$\pm$1.1 \\
          & $AC$[\%] & \textbf{23.0} & 7.3 & \textbf{27.0} & 35.7 & -0.3 \\
          & $Prec$[\%] & 38.9$\pm$1.9 & 14.6$\pm$7.7 & 43.0$\pm$5.1 & 56.5$\pm$5.8 & 15.7$\pm$0.5 \\
          & $Rec$[\%] & 58.0$\pm$11.6 & 65.3$\pm$27.6 & 80.3$\pm$10.9 & 55.1$\pm$1.6 & 62.8$\pm$10.3 \\
        \midrule
        \multirow{4}{*}{FedProx} & $F_1$[\%] & 42.4$\pm$6.3 & 22.5$\pm$4.7 & 52.8$\pm$6.9 & 58.0$\pm$6.2 & 25.6$\pm$0.9 \\
          & $AC$[\%] & 17.6 & 6.7 & 22.2 & 39.0 & 0.5 \\
          & $Prec$[\%] & 38.7$\pm$5.3 & 13.1$\pm$3.2 & 43.1$\pm$4.3 & 54.9$\pm$5.6 & 15.9$\pm$0.5 \\
          & $Rec$[\%] & 53.9$\pm$24.5 & 82.7$\pm$5.6 & 75.8$\pm$23.7 & 63.5$\pm$14.2 & 67.2$\pm$13.2 \\
        \midrule
        \multirow{4}{*}{MOON} & $F_1$[\%] & 32.2$\pm$7.2 & 17.6$\pm$1.9 & 51.2$\pm$3.3 & \textbf{61.6$\pm$2.2} & 26.2$\pm$0.9 \\
          & $AC$[\%] & 3.0 & 0.8 & 19.6 & \textbf{44.3} & 1.3 \\
          & $Prec$[\%] & 33.2$\pm$7.0 & 10.5$\pm$1.7 & 46.6$\pm$9.7 & 48.6$\pm$3.1 & 15.2$\pm$0.7 \\
          & $Rec$[\%] & 34.8$\pm$15.1 & 66.7$\pm$26.7 & 63.2$\pm$17.8 & 84.4$\pm$3.2 & 94.1$\pm$5.0 \\
        \midrule
        \multirow{4}{*}{TurboSVM} & $F_1$[\%] & 44.1$\pm$3.3 & 18.5$\pm$5.3 & 54.7$\pm$6.0 & 49.5$\pm$8.4 & 25.2$\pm$2.1 \\
          & $AC$[\%] & 20.0 & 1.9 & 25.4 & 26.7 & 0.0 \\
          & $Prec$[\%] & 36.5$\pm$4.8 & 10.7$\pm$4.0 & 46.2$\pm$5.0 & 53.3$\pm$6.2 & 15.6$\pm$1.2 \\
          & $Rec$[\%] & 59.1$\pm$14.7 & 77.3$\pm$13.0 & 73.0$\pm$25.5 & 48.1$\pm$14.5 & 66.5$\pm$10.3 \\
        \bottomrule
    \end{tabular}
    
    \label{tab:FL_SUM}
\end{table*}

Detecting mind wandering, engagement, and boredom solely from facial video is inherently challenging due to the subtle and often ambiguous nature of facial cues associated with these internal cognitive and affective states. Despite these challenges, our centralized learning model achieved performance levels that are in line with prior work in the centralized learning scenario, including studies by the original dataset publications. However, direct comparisons remain somewhat difficult due to differing train-test strategies (e.g., cross-validation vs. hold-out sets). For the \textbf{Colorado dataset}~\cite[][]{bosch2019automatic}, the original study reported an $F_1$ score of 42.1\% using deep neural networks applied to local binary pattern texture features extracted with OpenFace~\cite[][]{baltruvsaitis2016openface}. In a comparable setup, our bi-LSTM model achieved a slightly higher $F_1$ score of $43.9 \pm 1.4$ \%. Similarly, \citet[][]{anonim2023an} reported within-dataset $F_1$ scores of 45.9\% using explicit OpenFace features (e.g., action units and gaze) and 45.3\% using latent emotion-related features with a bi-LSTM model, which aligns well with our findings. For the \textbf{Korea dataset}~\cite[][]{lee2022predicting}, the authors achieved an $F_1$ score of $31 \pm 8$ \% on the ``not focused'' class using a deep neural network trained on 10-second time windows, incorporating features related to eye aspect ratio, emotion, gaze, and head movement. Using a different feature set but similar setting, our model yielded a slightly lower performance of $24.2 \pm 6.8$ \% $F_1$ score. In contrast, \citet[][]{anonim2023an} reported $F_1$ scores of 25.1\% using explicit features and 35.2\% using latent features in a cross-dataset prediction scenario on the same dataset. These improved performances underline the potential advantages of cross-dataset generalization in this domain. The authors of the multimodal \textbf{Germany dataset}~\cite[][]{buhler2024detecting} reported a best mind wandering $F_1$ score of 49.3\% using an XGBoost model trained solely on video features, which our approach surpassed by 0.7\% in the centralized and by 6.4\% in the federated learning scenario. However, their highest overall performance was achieved with a multimodal model combining all data sources, reaching a mind wandering $F_1$ score of 58.0\%, highlighting the advantages of leveraging multimodal information. In the publication introducing the \textbf{EngageNet dataset}~\cite[][]{singh2023have}, the authors reported classification accuracy across four graded levels of engagement. Using features extracted via OpenFace~\cite[][]{baltruvsaitis2016openface}, they achieved 62.37\% accuracy with an LSTM model based on eye gaze and head pose features. In comparison, our bi-LSTM model attained a slightly higher accuracy of $63.3 \pm 2.9$ \%. Similarly, \citet[][]{vedernikov2024tcct} reported 66.67\% engagement accuracy using an LSTM model with eye gaze, head pose, and facial action unit features, also extracted using OpenFace, which aligns with the performance observed in our study. For the \textbf{DAiSEE dataset}~\cite[][]{gupta2016daisee, kamath2016crowdsourced}, which includes four graded levels of boredom, the original authors reported a top-1 accuracy of 53.7\% using a long-term recurrent convolutional network~\cite[][]{donahue2015long} with five-fold cross-validation. Our bi-LSTM model achieved a comparable, though slightly lower, accuracy of $48.2 \pm 6.4$ \%. Since most subsequent studies have focused on engagement detection within DAiSEE, as summarized by \citet[][]{kumar2024measuring}, we limit our comparison to the results presented in the original publication.

On highly unbalanced and limited datasets with challenging classification tasks, bagging multiple models can sometimes perform worse than a single base model in terms of $F_1$ score (Table~\ref{tab:FL_SUM}), despite achieving higher accuracy (Table~\ref{tab:FL_SUM_additional}). In general cases, bagging employs different sampling strategies to create more balanced sub-datasets, however, in our case, the user-wise sampling preserves the class imbalance, as most participants contribute predominantly unbalanced samples. This, combined with the fact that bagging tends to further reduce variance and can overemphasize the majority class when aggregating predictions, leads to improved overall accuracy but poorer detection of the minority class.~\cite[][]{chen2024survey, feng2018class, blaszczynski2015neighbourhood}. Since the $F_1$ score focuses on the balance between precision and recall for the minority class, its value may decrease when the model favors majority class predictions to boost accuracy.

In general, federated learning is inferior to centralized learning regarding model performance, especially when data distribution among participating entities is non-iid~\cite[][]{fauzi2022comparative, nilsson2018performance}. However, in our experiments, the federated learning algorithms showed strong potential to outperform the centralized learning models on most datasets (Table~\ref{tab:FL_SUM}). Depending on the task and dataset, the best-performing approaches were FedAvg, FedAdam, FedAwS, and MOON.
A potential explanation for this phenomenon is that federated learning in our case provides benefits beyond what bagging achieved in the centralized setup. While bagging employs majority voting on the multiple models trained on individual participant data, it did not improve model performance in our experiments, likely because the sampling in bagging did not meaningfully alter the class imbalance or compensate for the data noise. In contrast, federated learning naturally partitions the data by participant, preserving individual-specific patterns and potentially reducing overfitting to dominant data characteristics. The aggregation of these diverse client models could act as a form of regularization that is better suited to the challenges of mind wandering detection, where data is noisy, imbalanced, and highly individual. 
A further possible reason for federated learning outperforming ensemble bagging in our benchmark lies in that compared to majority vote in bagging, the averaging-based model aggregation in federated learning has a better potential to profit from model personalization~\cite{tan2022towards, mansour2020three, zhang2023fedala}. In this regard, we provide a toy example for model averaging exceeding majority vote in the Appendix~\ref{sec:bagging}. The purpose of this example is not to serve as strict proof but as a heuristic.

We also dived into the model training process. The training loss for all five datasets remained at a higher value in the centralized setting compared to FedAvg for example. While the averaged training loss curve appeared smoother for centralized learning, FedAvg consistently achieved lower final loss, indicating better convergence. Additionally, centralized learning exhibited a larger standard deviation (Table~\ref{tab:FL_SUM}), suggesting that while it occasionally outperformed federated learning algorithms for certain random seeds, its performance was less stable. This variability can be explained by factors such as the sensitivity of centralized learning to weight initialization, whereas federated learning benefits from iterative local updates across clients, leading to more robust and consistent optimization.
Lastly, our paper is not the first to show that federated learning can outperform centralized learning, namely, \citet[][]{li2024fedbchain} demonstrates that federated learning, utilizing a single-layer DeepConvLSTM architecture, outperforms centralized training methods in precision, recall, and $F_1$ score across multiple datasets and federated strategies. \citet[][]{schwinn2024comparative} demonstrates that federated learning can achieve comparable or slightly improved performance compared to centralized learning approaches, while preserving patient data privacy and offering significant benefits for hospitals with limited datasets.

\subsection{Ablation Study}
\subsubsection{Feature Selection}

\begin{table*}[tb]
    \centering
    \scriptsize
    \caption{ Model performance in centralized and decentralized learning (FedAvg~\cite[][]{mcmahan2017communication}) scenarios trained on a variety of feature sets (from EmoNet~\cite[][]{toisoul2021estimation} and OpenFace gaze~\cite[][]{baltrusaitis2018openface}) on EngageNet~\cite[][]{singh2023have}, Colorado~\cite[][]{bosch2019automatic}, Korea~\cite[][]{lee2022predicting} and DAiSEE~\cite[][]{gupta2016daisee, kamath2016crowdsourced} datasets. All results are obtained with bi-LSTM model.}
    \begin{tabular}{p{8mm} p{8mm} p{7mm} p{14mm} p{14mm} p{14mm} p{14mm} p{14mm}}
    \toprule
        Features &\multicolumn{2}{c}{Performance} & Colorado & Korea & Germany & EngageNet & DAiSEE \\
        \midrule
        \multirow{6}{*}{\begin{sideways} \thead{EmoNet} \end{sideways}} & \multirow{3}{*}{\thead{non-FL \\ bi-LSTM}} & $F_1$[\%] & 39.4$\pm$9.8 & 75.8$\pm$10.1 & 43.5$\pm$12.4 & 65.0$\pm$4.1 & 56.1$\pm$7.4 \\
         & & $Prec$[\%] & 61.4$\pm$5.0 & 91.0$\pm$2.6 & 41.0$\pm$9.4 & 66.3$\pm$5.3 & 77.0$\pm$1.2\\
         & & $Rec$[\%] & 41.4$\pm$6.8 & 68.3$\pm$12.7 & 48.9$\pm$21.5 & 66.1$\pm$4.5 & 49.5$\pm$7.1\\
         \cmidrule{2-8}
        & \multirow{3}{*}{FedAvg} & $F_1$[\%] & 63.8$\pm$4.7 & 84.7$\pm$5.3 & 40.8$\pm$7.9 & 66.9$\pm$4.6 & 56.9$\pm$7.4 \\
         & & $Prec$[\%] & 64.9$\pm$4.1 & 88.7$\pm$1.8 & 47.8$\pm$8.2 & 69.0$\pm$3.4 & 76.2$\pm$1.1\\
         & & $Rec$[\%] & 64.3$\pm$6.0 & 82.1$\pm$9.4 & 41.6$\pm$20.2 & 66.9$\pm$4.8 & 50.2$\pm$7.8\\
        \midrule
        \multirow{6}{*}{\begin{sideways} \thead{OpenFace}\end{sideways}}  & \multirow{3}{*}{\thead{non-FL \\ bi-LSTM}} & $F_1$[\%] & 46.5$\pm$8.2 & 16.2$\pm$18.3 & 35.2$\pm$22.5 & 73.0$\pm$2.3 & 55.7$\pm$1.6 \\
         & & $Prec$[\%] & 58.7$\pm$4.8 & 74.6$\pm$4.2 & 62.7$\pm$22.4 & 74.2$\pm$3.4 & 74.4$\pm$1.1\\
         & & $Rec$[\%] & 48.2$\pm$13.0 & 15.6$\pm$12.3 & 35.8$\pm$32.0 & 74.2$\pm$2.7 & 50.4$\pm$14.4\\
         \cmidrule{2-8}
        & \multirow{3}{*}{FedAvg} & $F_1$[\%] & 60.5$\pm$2.2 & 51.9$\pm$13.5 & 23.3$\pm$8.6 & 70.9$\pm$4.9 & 66.5$\pm$6.3 \\
         & & $Prec$[\%] & 58.8$\pm$6.4 & 93.7$\pm$0.7 & 64.3$\pm$20.0 & 73.0$\pm$6.6 & 75.2$\pm$1.4\\
         & & $Rec$[\%] & 69.0$\pm$2.1 & 41.9$\pm$12.6 & 15.3$\pm$6.8 & 70.4$\pm$7.2 & 61.9$\pm$8.7\\
        \midrule
        \multirow{6}{*}{ \begin{sideways} \thead{OpenFace \\ gaze}\end{sideways}} & \multirow{3}{*}{\thead{non-FL \\ bi-LSTM}} & $F_1$[\%] & 54.4$\pm$14.3 & 72.3$\pm$13.1 & 37.0$\pm$7.0 & 77.8$\pm$2.6 & 44.2$\pm$18.2 \\
         & & $Prec$[\%] & 72.7$\pm$4.0 & 88.9$\pm$1.9 & 45.1$\pm$5.3 & 79.5$\pm$0.7 & 76.8$\pm$1.6\\
         & & $Rec$[\%] & 58.0$\pm$14.8 & 64.7$\pm$17.4 & 33.7$\pm$12.1 & 78.2$\pm$1.7 & 40.5$\pm$15.7\\
         \cmidrule{2-8}
        & \multirow{3}{*}{FedAvg} & $F_1$[\%] & 71.7$\pm$0.4 & 77.1$\pm$14.5 & 36.3$\pm$21.7 & 75.0$\pm$0.7 & 79.0$\pm$0.4 \\
         & & $Prec$[\%] & 71.5$\pm$0.4 & 88.3$\pm$1.8 & 34.0$\pm$19.1 & 76.8$\pm$1.8 & 74.6$\pm$3.4\\
         & & $Rec$[\%] & 72.7$\pm$1.1 & 72.0$\pm$19.2 & 44.2$\pm$36.2 & 76.2$\pm$1.0 & 85.5$\pm$0.0\\
        \midrule
        \multirow{6}{*}{ \begin{sideways} \thead{ EmoNet+ \\ OpenFace} \end{sideways}}  & \multirow{3}{*}{\thead{non-FL \\ bi-LSTM}} & $F_1$[\%] & 43.2$\pm$12.9 & 66.0$\pm$15.0 & 40.9$\pm$12.1 & 68.9$\pm$1.7 & 56.1$\pm$4.0 \\
         & & $Prec$[\%] & 62.9$\pm$7.0 & 91.2$\pm$2.5 & 45.6$\pm$3.5 & 70.5$\pm$1.4 & 76.4$\pm$0.5\\
         & & $Rec$[\%] & 44.6$\pm$9.9 & 56.6$\pm$16.3 & 40.0$\pm$16.8 & 70.9$\pm$1.2 & 49.1$\pm$4.3\\
         \cmidrule{2-8}
        & \multirow{3}{*}{FedAvg} & $F_1$[\%] & 61.8$\pm$1.5 & 80.6$\pm$7.9 & 46.4$\pm$23.2 & 66.6$\pm$0.3 & 56.8$\pm$14.8 \\
         & & $Prec$[\%] & 62.3$\pm$1.9 & 88.4$\pm$1.0 & 55.2$\pm$25.4 & 67.8$\pm$1.3 & 75.9$\pm$1.0\\
         & & $Rec$[\%] & 61.9$\pm$2.3 & 75.6$\pm$11.7 & 65.8$\pm$35.9 & 68.6$\pm$8.6 & 51.4$\pm$13.3\\
        \bottomrule
    \end{tabular}
    
    \label{tab:ablation}
\end{table*}

To test the robustness and sensitivity of the centralized and decentralized frameworks to different feature sets (Table~\ref{tab:ablation}), we conducted a series of experiments, where we selected a variety of input features. In almost all cases, FedAvg~\cite[][]{mcmahan2017communication} achieved better performance compared to a centralized setting trained on the same feature sets. Models trained only on the OpenFace~\cite[][]{baltrusaitis2018openface} eight gaze features resulted in high model performance. However, upon closer examination, it became apparent that the model failed to learn effectively, as indicated by a training MCC score close to zero. Consequently, it can be inferred that these models predicted the majority class. By combining all features from the EmoNet~\cite[][]{toisoul2021estimation} and OpenFace~\cite[][]{baltrusaitis2018openface}, the input dimension of each sample increased from $124 \times 520$ to $124 \times 1221$, which increased the computational time and overloaded the model, and hence resulted in decreased model performance. Overall, the model performances in centralized and decentralized settings have only small oscillations (excluding the model trained on the OpenFace~\cite[][]{baltrusaitis2018openface} gaze features), underscoring the robustness and generalizability of our models.

\subsubsection{Illumination}
\begin{figure}[t]
	\centering

	\subfloat[Frame from the original video.]{\includegraphics[height=3cm, keepaspectratio]{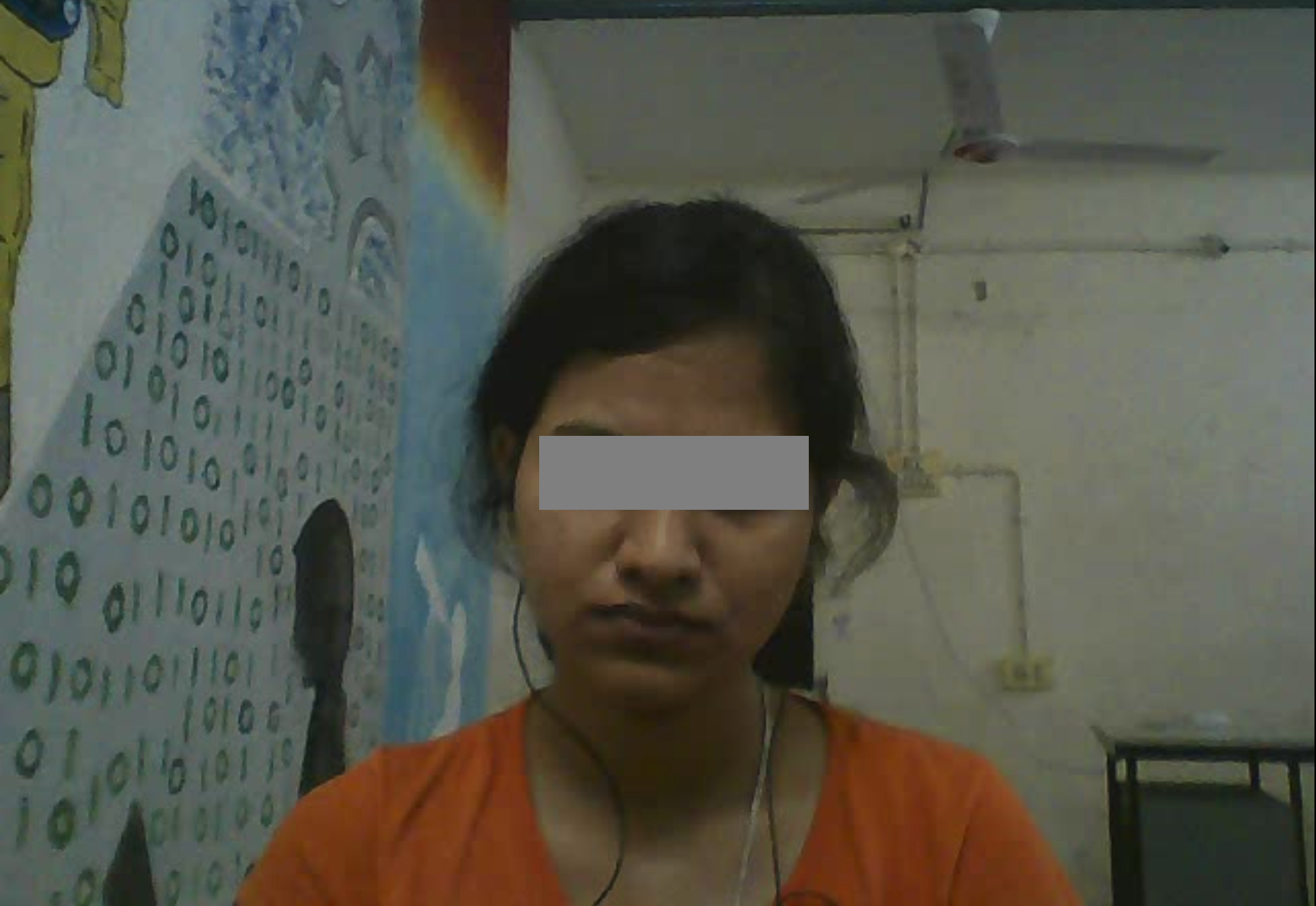}
	\label{fig:orig_ill}}
\hspace{1cm}
	\subfloat[Frame from the enhanced video.]{\includegraphics[height=3cm, keepaspectratio]{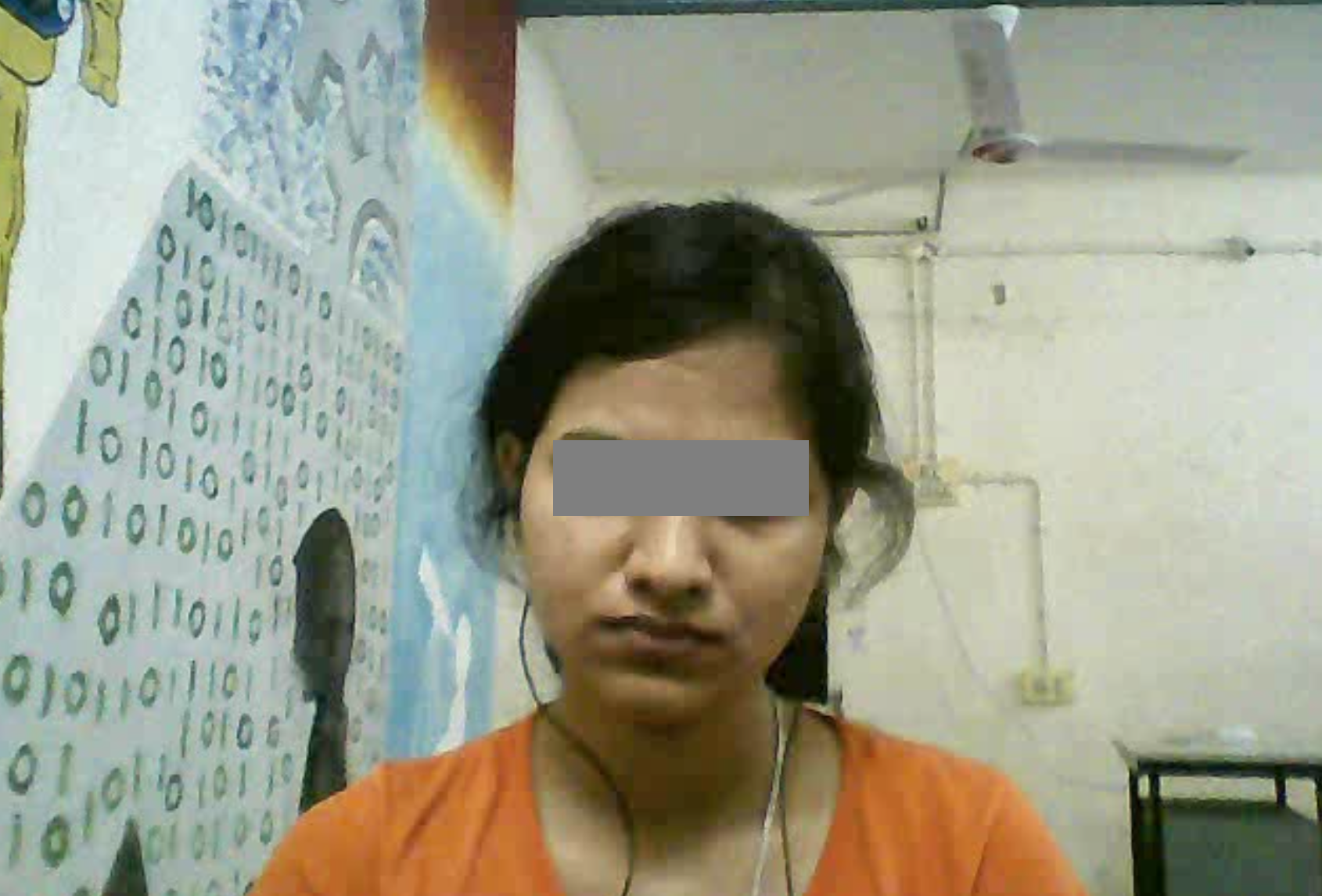}
	\label{fig:ill_enhanced}}
	\hfil
	 \caption{Before and after applying video enhancement from Harmonizer~\cite[][]{ke2022harmonizer} on a video from DAiSEE~\cite[][]{gupta2016daisee, kamath2016crowdsourced}.}
    \label{fig:illumination}
\end{figure}

In the case of datasets (~\cite[][]{lee2022predicting, singh2023have, gupta2016daisee, kamath2016crowdsourced}), where videos were recorded in diverse environments, the illumination also varies on a larger scale. For the students who recorded themselves in a darker environment, the recognition of disengagement could become challenging. To solve this problem, we first filtered out the videos with low brightness by converting the video frames into grayscale and checking for frames with an average pixel value of less than a hundred. If low brightness persists for a
continuous one-second duration, the entire video is identified as inadequately lit. In this sense, comparing all five datasets we have used, the Colorado~\cite[][]{bosch2019automatic} and Germany~\cite[][]{buhler2024detecting} datasets were recorded in the lab and hence has high-quality frames, while the other three datasets showed different levels of illumination problems: 1 video in the Korea~\cite[][]{lee2022predicting}, 352 videos in the EngageNet~\cite[][]{singh2023have}, and 495 videos in the DAiSEE~\cite[][]{gupta2016daisee, kamath2016crowdsourced} dataset were detected via our pipeline. This indicates a good video quality in the Korea~\cite[][]{lee2022predicting} dataset, thus illumination enhancement is not necessary. Therefore, we only conducted experiments on the EngageNet~\cite[][]{singh2023have} and DAiSEE~\cite[][]{gupta2016daisee, kamath2016crowdsourced} datasets. We used the open-source tool, Harmonizer~\cite[][]{ke2022harmonizer}, to increase the illumination (Figure~\ref{fig:illumination}). Then, the EmoNet~\cite[][]{toisoul2021estimation} and OpenFace~\cite[][]{baltrusaitis2018openface} features were re-extracted. The illumination enhancement resulted in higher model performance (Table~\ref{tab:ablation_illumination}), however, the results are in the same range for all settings.

\begin{table*}[t]
    \centering
    \caption{ Model performance in centralized and decentralized learning (FedAvg) scenarios trained on the original and enhanced illumination settings on EngageNet~\cite[][]{singh2023have} and DAiSEE~\cite[][]{gupta2016daisee, kamath2016crowdsourced}  datasets. All results are obtained with bi-LSTM model.}
    \begin{tabular}{c c c c c}
    \toprule
         &\multicolumn{2}{c}{Performance} & EngageNet & DAiSEE \\
        \midrule
        \multirow{6}{*}{Original videos} & \multirow{3}{*}{non-FL (bi-LSTM)} & $F_1$[\%] & 51.4$\pm$6.0 & 25.7$\pm$0.6  \\
         & & $Prec$[\%] & 55.7$\pm$2.3 & 16.3$\pm$0.7 \\
         & & $Rec$[\%] & 48.6$\pm$10.0 & 61.6$\pm$6.8 \\
        & \multirow{3}{*}{FedAvg} & $F_1$[\%] & 57.4$\pm$4.6 & \textbf{26.2$\pm$1.8} \\
        & & $Prec$[\%] & 51.9$\pm$3.1 & 16.3$\pm$1.4 \\
         & & $Rec$[\%] & 65.3$\pm$10.4 & 67.3$\pm$6.8 \\
        \midrule
        \multirow{6}{*}{ \begin{tabular}{@{}c@{}}Illumination \\ improved videos\end{tabular}}  & \multirow{3}{*}{non-FL (bi-LSTM)} & $F_1$[\%] & 51.4$\pm$4.8 & \textbf{26.2$\pm$0.5} \\
         & & $Prec$[\%] & 60.3$\pm$9.7 & 16.4$\pm$0.2 \\
         & & $Rec$[\%] & 48.0$\pm$14.1 & 65.9$\pm$5.2 \\
        & \multirow{3}{*}{FedAvg} & $F_1$[\%] & \textbf{59.0$\pm$4.5} & \textbf{26.2$\pm$1.1} \\
        & & $Prec$[\%] & 55.8$\pm$9.0 & 15.7$\pm$0.9 \\
         & & $Rec$[\%] & 66.0$\pm$12.6 & 81.6$\pm$14.8 \\
        \bottomrule
    \end{tabular}
    
    \label{tab:ablation_illumination}
\end{table*}

\section{Discussion}
\subsection{General Insights}
In this work, we employed federated learning to detect multiple manifestations of disengagement during online learning. We benchmarked six recent federated learning algorithms that aim to boost convergence or address data heterogeneity. By leveraging the federated learning paradigm, we not only offered a foundational guarantee of user privacy due to its inherent design but also improved model performance compared to traditional centralized learning. Furthermore, we addressed the challenge for participants with glasses and improved model performance by incorporating glass-related features for both those with and without glasses across two datasets.

Our models achieved performance levels consistent with prior studies across all datasets, confirming that our centralized learning results align with established benchmarks in the field. These findings underscore both the difficulty of the task and the need for multimodal data or alternative approaches to further improve detection accuracy.These results indicate that, across all examined datasets, the attainable model performance is considerably lower than that typically observed on benchmark image classification datasets such as CIFAR-10~\cite[][]{krizhevsky2009learning}. While the results remain above chance, confirming that meaningful features can be extracted from facial behavior, we acknowledge that current performance levels may still produce false positives if deployed in real-world educational settings, potentially distracting rather than supporting learners. Notably, we demonstrated that our centralized learning results align well with prior findings reported in the original dataset publications, thereby validating our baseline. Moreover, in four out of five datasets, our federated learning models outperformed their centralized counterparts, highlighting the potential of federated approaches to improve stability and performance in challenging, unbalanced, user-independent scenarios. Finally, the limited dataset size (particularly for the Korea dataset) further constrained model generalizability, emphasizing the importance of larger and more diverse training data to enhance robustness and reliability in future applications.

\subsection{Limitations}
By incorporating glass-related features, we improved model performance for participants with glasses and without across two datasets. Our results indicate that MeGlass features provided slight performance gains for non-mind-wandering datasets, but offered no consistent benefit for mind-wandering detection, where baseline models often performed similarly or better. These findings should be interpreted cautiously, as the limited and imbalanced representation of participants with glasses in our datasets likely constrained the ability to generalize the value of these features. Our results disclose that a facial landmark detection model and an eye-tracking model that is more robust against eyeglass reflection are desired. Moreover, we did not observe any enhancement concerning the Korea dataset~\cite[][]{lee2022predicting}. This could be attributed to the limited size of the dataset (only 15 participants), which makes it exceedingly challenging for the model to leverage additional features due to the difficulty in generalizability based on the low number of participants included in the dataset. 

An interesting phenomenon that could influence the predictions is the Hawthorne effect~\cite[][]{sedgwick2015understanding}, which points to a changed learning behavior due to the fact that the students know that they are being observed and change their natural behavior accordingly. Besides, a high variety in the model performance on each dataset can be observed, which can be attributed to the different behavior of participants in diverse environments: the students who were watching lectures at home would behave less formally than those in a laboratory for the experiments. This difference over datasets can also be explained by recognizing that albeit mind wandering, disengagement, and boredom are related to each other and can have similar negative effects on the learning outcome, they are distinct entities, entailing very distinct observable indicators, and their detection difficulties can differ accordingly. For example, furrowed brows, wandering gazes, or restless movements could point to disengagement~\cite[][]{firat2023predicting}, while mind wandering detection recognition from facial and behavioral cues is a challenging task for humans~\cite[][]{bosch2019automatic}. It is also worth mentioning that labeling such datasets is extremely challenging~\cite[][]{bosch2022can}, and the labeling method can also differ between questionnaires~\cite[][]{singh2023have} and direct questions~\cite[][]{lee2022predicting}, self-reports~\cite[][]{bosch2019automatic}, and annotations~\cite[][]{gupta2016daisee, kamath2016crowdsourced}. These phenomena could also explain the difference in experimental results achieved on the different datasets. Therefore, more standardized data collection pipelines and larger datasets in this field are desired.

Among the five datasets employed for our experiments, some are already recorded outside a laboratory, which makes it possible to transfer our results to real word applications. The datasets show significant imbalance, both in the ratio of positive to negative samples and in the distribution between participants with and without glasses, closely reflecting real-life scenarios. Moreover, the number of samples attributed to each individual in the datasets varied widely, from participants with less than 5 samples to more than 190 samples (Figure~\ref{fig:histo1}), mirroring the dynamics of real-life scenarios where certain users may have limited samples. Additionally, in each training round within federated learning, only half of the users actively took part in the training, simulating participants who were not consistently present at their laptops. These assumptions were deliberately made to realistically model scenarios encountered in everyday life, such as helping to fill the gaps with missed information when mind wandering is detected, giving alerts to divers and healthcare professionals to keep them focused, or to increase mental well-being ~\cite[][]{kuvar2023detecting}. Another advantage of our federated learning approach is that when more and more students would provide their data, our models could be used for further training. By counting on potential data collection, our models would achieve better model performance and could increase the level of generalization.
\begin{figure}[t]
	\centering

	\subfloat[Hidden/Occluded face.]{\includegraphics[height=2.9cm, keepaspectratio]{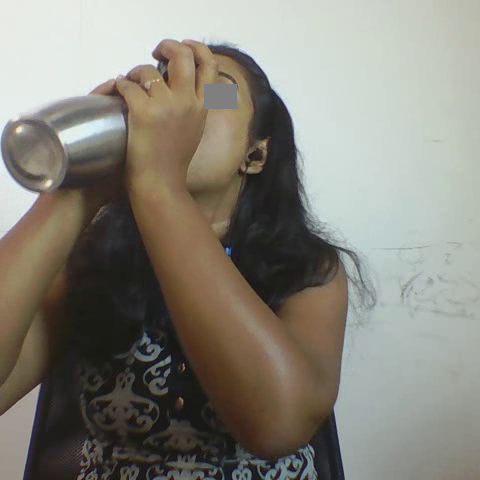}
	\label{fig:drink}}
 \hspace{1cm}
	\subfloat[Suboptimal body posture.]{\includegraphics[height=2.9cm, keepaspectratio]{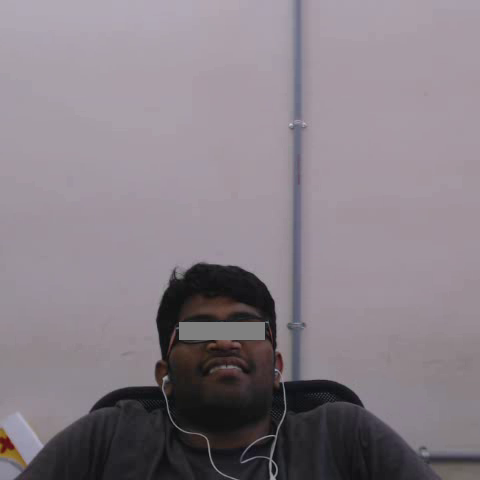}
	\label{fig:bad_setting}}
 \hspace{1cm}
	\subfloat[Hands covering one eye.]{\includegraphics[height=2.9cm, keepaspectratio]{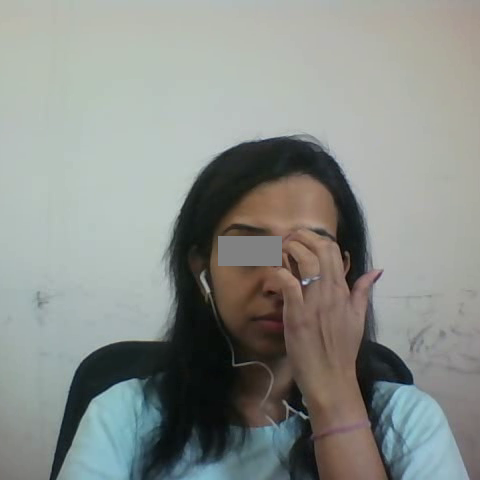}
	\label{fig:hands}}
	\hfil
 \caption{Examples from the DAiSEE~\cite[][]{gupta2016daisee, kamath2016crowdsourced} dataset which demonstrate some common scenarios where face recognition is challenging.}	
    \label{fig:limitation}
\end{figure}

\subsection{Ethical Considerations}
With AI increasingly employed in educational contexts, ethical considerations are of utmost importance~\cite[][]{FUTTERER2025100483}. Our approach does not perform explicit emotion recognition but rather utilizes latent features extracted from a pre-trained, publicly available emotion detection network~\cite[][]{toisoul2021estimation}, ensuring that no direct emotion classification is conducted. This distinction is crucial in light of the European Union's Artificial Intelligence (AI) Act~\cite[][]{AIAct}, which imposes strict limitations on biometric data processing, particularly in educational contexts. Moreover, our focus on glasses stems from the technical challenge posed by reflections, which can impact model performance. Importantly, our federated learning approach ensures that video data remains entirely on learners’ devices, eliminating the need for centralized data collection and thereby strengthening privacy protection. This decentralized approach aligns with the General Data Protection Regulation of the European Union~\cite[][]{drachsler2016privacy} data minimization principles and the AI Act’s~\cite[][]{AIAct} emphasis on privacy-preserving AI. Additionally, since our method can be executed on standard devices without requiring specialized hardware, it democratizes the use of AI for educational technologies, promoting equitable adoption across different learning environments.

The use of facial videos in online learning environments to assess student engagement raises significant ethical questions, especially concerning privacy, autonomy, and informed consent. Therefore, it is essential that students maintain control over their personal data and their level of participation. Real-time data processing raises concerns about informed consent and the extent to which students can opt in or out without pressure~\cite[][]{rubel2016student, prinsloo2015student}. By processing video data locally without transmission to external servers, our approach preserves privacy while still allowing models to improve over time. Additionally, we do not process the raw video data; instead, we provide the models with only the extracted facial and emotion-related features, further enhancing user data privacy. Institutions have a responsibility to ensure transparency and ethical handling of student data to foster trust in AI-driven educational tools~\cite[][]{jones2020matter}. This privacy-preserving design mitigates risks associated with data collection to a significant extent while still enabling real-time adaptation and potential on-device model updates to improve personalization.

Regarding data privacy, a further concern is that federated learning introduces a potential vulnerability to privacy breaches in the event of a cyberattack both on the client and server side, which requires further discussion in the future. Recently, a promising trend in the federated learning domain is to improve the security level and privacy protection through the integration of other techniques, such as by utilizing blockchains~\cite[][]{9170559, wang2021blockchainbased, 9184854}, applying differential privacy~\cite[][]{mcmahan2017learning, wei2019federated}, or using secure aggregation~\cite[][]{bonawitz2017practical}, and the application of these methods in remote learner state detection remains as future work. Another potential direction is personalized federated learning~\cite[][]{tan2022towards}, which can be beneficial both for addressing data heterogeneity while preserving data privacy. 

\subsection{Future Work}
A further direction of follow-up work is the feedback-triggered software detecting disengagement and helping the participants refocus their attention. To develop effective educational solutions that deploy our trained model, better model performance is needed since false predictions could distract or even irritate the users. Detecting scenarios when the faces of the students are invisible or hard to recognize (Figure~\ref{fig:limitation}) and disabling the inferences for those particular times would increase the model performance. As an initial prototype, utilization of an online video-watching platform akin to crowdsourcing platforms such as Prolific~\cite[][]{palan2018prolific} is feasible. Here, the background processes would encompass feature extraction and the deployment of our model. In the event of a mind wandering episode, an intervention would promptly emerge, and an adept assistant could be integrated to summarize the preceding two minutes of the lecture, aiding participants in comprehending any missed content. Alternatively, a question prompt might pop up, enabling participants to seek clarification from different sources, such as a large language model. Such an application holds the potential to realign participants' focus and bridge any knowledge gaps, ensuring they stay abreast with the ongoing lecture. Besides, with such an application the collection of new data would also be possible, where participants could report mind wandering, or professionals could determine the occurrence of disengagement or boredom. By collecting data from participants with more diverse demographic and geographic backgrounds in such web-based experiments, training of more inclusive models for different participant types would be possible. In contrast, most existing datasets were captured in laboratories; thus, the samples come from small and locally recruited participants, which might result in non-generalizable trained models~\cite[][]{arnett2016neglected, henrich2010weirdest, reinecke2015labinthewild}.

\section{Conclusion}
In this work, we focused on identifying remote learner states within the context of online learning. Our findings suggest that while MeGlass features slightly improved performance on non-mind-wandering datasets, they did not consistently enhance mind-wandering detection, likely due to limited representation of participants with glasses in the data. This was achieved by adding a feature vector from a specialized neural network trained to recognize eyewear. Subsequently, we extended our investigation to embrace a decentralized learning scenario using federated learning techniques, enhancing data privacy. We achieved higher accuracy and $F_1$ score values by applying federated learning, based on our extensive experiments on five datasets. The novelty of our work is the privacy-by-design federated learning algorithms applied in the educational technologies domain, where the focus lies on the immediate guidance of learners during online lectures. Our proposed framework holds promise in augmenting the efficacy of online learning in a privacy-preserving fashion, particularly for disengaged or bored students.

\section*{Statements and Declarations}
\subsection*{Funding Declaration}
We acknowledge the funding by the Deutsche Forschungsgemeinschaft (DFG, German Research Foundation) Project number 491966293.

\subsection*{Competing Interests}
The authors declare that they have no competing interests.

\subsection*{Data Statement}
We publish our implementation (\url{https://gitlab.lrz.de/hctl/digital-self-control}) for reproducibility.

\newpage

\bibliography{sn-bibliography}

\clearpage
\begin{appendices}


\section{Data Distribution}
Details about data distribution of all five datasets after data preprocessing are given in Table~\ref{tab:dataset_split_details} and \ref{tab:dataset_split_details_test}. A slight variation in the distribution between centralized and decentralized settings is observed in the datasets, which can be attributed to the reduced number of participants with glasses. To create the training, validation, and test sets, we employed random sampling. Upon noticing the uneven distribution of participants with and without glasses, we made small modifications to the non-FL case where we tested the MeGlass~\cite[][]{guo2018face} features. However, we kept the setups unchanged for the federated learning scenarios.

\begin{table*}[htb]
    \centering
    \caption{Details about the exact number of positive (disengagement, mind wandering, and boredom) and negative samples, number of participants with and without glasses, and the gender of the participants in the train set in each dataset on EngageNet~\cite[][]{singh2023have}, Colorado~\cite[][]{bosch2019automatic}, Korea~\cite[][]{lee2022predicting} and DAiSEE~\cite[][]{gupta2016daisee, kamath2016crowdsourced}).}
    \begin{tabular}{c c c c c c c}
    \toprule
         & \multicolumn{6}{c}{Train and validation sets}   \\
        & Dataset & samples & positive & users & glasses  & male \\
        \midrule
        \multirow{5}{*}{\begin{sideways} non-FL \end{sideways}} & Colorado & 3001 & 895 & 117 & 9 & 46  \\
        & Korea & 979 & 170 & 12 & 6  & 5  \\
        & Germany & 846 & 317 & 69 & 22 & 13 \\
        & EngageNet & 8193 & 2495 & 83 & 29 & 58  \\
        & DAiSEE & 7835 & 2095 & 100 & 46 & 73 \\
        \midrule
        \multirow{5}{*}{\begin{sideways} FL \end{sideways}} & Colorado & 2998 & 905 & 117 & 8 & 45  \\
        & Korea & 979 & 191 & 12 & 7 & 6  \\
        & Germany & 846 & 317 & 69 & 22 & 13 \\
        & EngageNet & 8193 & 2495 & 83 & 29 & 58 \\
        & DAiSEE & 7835 & 2095 & 100 & 46 & 73 \\
        \bottomrule
        
    \end{tabular}
    
    \label{tab:dataset_split_details}
    
\end{table*}

\begin{table*}[htb]
    \centering
    \caption{Details about the exact number of positive (disengagement, mind wandering, and boredom) and negative samples, the number of participants with and without glasses, and the gender of the participants in the  test set. \textit{M} and \textit{F} are referring to the number of male and female participants in each dataset on EngageNet~\cite[][]{singh2023have}, Colorado~\cite[][]{bosch2019automatic}, Korea~\cite[][]{lee2022predicting} and DAiSEE~\cite[][]{gupta2016daisee, kamath2016crowdsourced}).}
    \begin{tabular}{c c c c c c c}
    \toprule
        &  & \multicolumn{5}{c}{Test Set}  \\
        & Dataset & Samples & Positive & Users & Glasses & Male \\
        \midrule
        \multirow{5}{*}{\begin{sideways} non-FL \end{sideways}} & Colorado  & 302 & 100 & 13 & 3 & 4 \\
        & Korea & 241 & 36 & 3 & 1 & 2  \\
        & Germany & 93 & 38 & 8 & 2 & 2    \\
        & EngageNet & 811 & 308 & 16 & 4 & 7 \\
        & DAiSEE & 1090 & 158 & 12 & 8 & 8 \\
        \midrule
        \multirow{5}{*}{\begin{sideways} FL \end{sideways}} & Colorado & 305 & 90 & 13 & 4 & 5 \\
        & Korea & 241 & 15 & 3 & 0 & 1 \\
        & Germany & 93 & 38 & 8 & 2 & 2    \\
        & EngageNet  & 811 & 308 & 16 & 4 & 7 \\
        & DAiSEE & 1090 & 158 & 12 & 8 & 8 \\
        \bottomrule
        
    \end{tabular}
    
    \label{tab:dataset_split_details_test}
    
\end{table*}

\section{Data Preprocessing}
As video quality varies among different samples to a large extent, we first discarded some samples. Specifically, videos were omitted from consideration in cases if the OpenFace~\cite[][]{baltrusaitis2018openface} failed to capture facial landmarks continuously for ten consecutive frames, which may occur due to situations like accidental face scratching. Additionally, if the framework failed to recognize the face for a total duration exceeding thirty frames, it was also excluded. In instances where the face was unrecognizable for fewer than ten frames, we applied a corrective measure by discarding the problematic frames and substituting them with replicating the last successfully identified frame. Face recognition might falter due to various factors, such as participants tilting their heads, moving outside the video frame, or positioning in a manner where only a segment of their face is visible (like just the forehead). Additionally, inadequate lighting on their faces can also hinder recognition. After these data preprocessing steps, 6 and 12 participants with 19 and 48 positive and 42 and 111 negative labels are removed from the EngageNet~\cite[][]{singh2023have} and Colorado~\cite[][]{bosch2019automatic} datasets, respectively. 

To have comparable results with a trained MLP model, we simplified the dimension of the input feature vector by averaging every ten features corresponding to single frames into one and cutting the last four frames. The input dimension was changed from $124 \times 520$ to $12 \times 520$. As a next step, we flattened this feature matrix into a vector.

\section{Facial Features}\label{app:facial_features}
Consistent with numerous related works~\cite[][]{reichle2012using, brishtel2020mind, dewan2019engagement, angeline2021review, kim2018detecting, krithika2016student, hutt2021breaking, d2016attending}, forms of disengagement have been linked to facial expressions, eye gaze patterns, and human emotions. Given this association, we extracted these features for the classification of all our video samples. Instead of using raw pixel values from each frame, we emphasized the significance of features by leveraging those extracted by trained neural networks or from facial landmark coordinates. This approach not only simplifies our model's input but also underscores the pivotal role of feature quality in detecting self-regulation problems. Hence, evaluating the feature extractor becomes crucial whenever feasible. Many previous works~\cite[][]{bosch2019automatic, singh2023have} use features generated by the OpenFace frameworks~\cite[][]{baltruvsaitis2016openface, baltrusaitis2018openface}, which are specialized for facial behavior and eye tracking and retrieves explicit information like facial landmarks, facial action units, head pose and gaze direction in a vector of dimension 709. Since the predicted facial landmark coordinates are measurable in contrast to extracted feature vectors from a neural network, the quality of the predicted points can be evaluated on datasets containing manual labels describing facial landmark coordinates. An example of such a dataset is the Eyeblink8~\cite[][]{drutarovsky2014eye} dataset, which contains ground truth eye corner coordinates. Since the correct detection of the eyes plays a crucial role in disengagement detection, we only focus on evaluating these predicted coordinates by OpenFace~\cite[][]{baltrusaitis2018openface}.

\begin{table}[htb]
    \centering
    \caption{OpenFace~\cite[][]{baltrusaitis2018openface} prediction quality evaluation on the Eyeblink8~\cite[][]{drutarovsky2014eye} dataset. The absolute distance is measured from the ground truth location. The $> number $ refers to the number of frames, where the absolute distance between the OpenFace prediction and the ground truth value is greater than the \textit{number}.}
    \begin{tabular}{ p{15mm} p{10mm} p{10mm} p{10mm} p{10mm} p{10mm} p{10mm} p{10mm} p{10mm} p{9mm}}
    \toprule
         & \multicolumn{2}{c}{Client 1} & \multicolumn{2}{c}{Client 2} & \multicolumn{2}{c}{Client 3}& \multicolumn{2}{c}{Client 4}\\
         & video 1 & video 2 & video 1 & video 2 & video 1 & video 2 & video 1 & video 2 \\
         Glasses? & no & no & no & no & no & no & no & yes \\
         \midrule
        abs. dist. & 2.99 & 3.38 & 2.56 & 1.14 & 1.01 & 0.75 & 1.04 & 1.18 \\
        $>10$  & 397.25 & 362.63 & 215.75 & 13.50 & 4.25 & 0.00 & 0.00 & 0.00 \\
        $>20$  & 75.75 & 123.88 & 74.00 & 3.00 & 0.00 & 0.00 & 0.00 & 0.00 \\
        $>50$  & 4.25 & 7.50 & 1.50 & 3.00 & 0.00 & 0.00 & 0.00 & 0.00 \\
        \midrule
        \# of frames & 15711 & 11123 & 9216 & 5315 & 10663 & 5060 & 9014 & 4890 \\
        avg. conf. & 0.978 & 0.979 & 0.979 & 0.974 & 0.980 & 0.980 & 0.980 & 0.972 \\
        \bottomrule
    \end{tabular}
    
    \label{tab:of_quality}
\end{table}

In the Eyeblink8~\cite[][]{drutarovsky2014eye} dataset, the eye corner coordinates are labeled, which makes it possible to calculate the distance between the eye corner coordinates predicted by OpenFace~\cite[][]{baltrusaitis2018openface} and the given ones. This dataset contains students sitting at home, acting naturally, in the same setup as the above-mentioned five datasets~\cite[][]{singh2023have, bosch2019automatic, lee2022predicting, gupta2016daisee, kamath2016crowdsourced}. The Eyeblink8~\cite[][]{drutarovsky2014eye} dataset contains 8 videos of 4 individuals recorded by web cameras (resolution: $640\times 480$). For each frame, the left and right eye coordinates are given pixel-wise. In Table~\ref{tab:of_quality}, the quality of OpenFace~\cite[][]{baltrusaitis2018openface} feature extraction is provided. The distance between the predicted 4 eye coordinates and the ground truth (GT) coordinates is compared on each frame. Overall, the confidence of the feature extraction for the whole video provided by OpenFace~\cite[][]{baltrusaitis2018openface} is above 0.97 for all videos, which generally would point to successful feature extraction on almost all frames. On the other hand, three out of eight videos contain more than 200 incorrect predictions (more than a ten-pixel difference between the predicted and ground truth of the coordinates), while the remaining only consist of almost perfect predictions. This suggests that although the extracted OpenFace~\cite[][]{baltrusaitis2018openface} features have a decent quality in general, for some participants, it can make incorrect predictions without a drop in its confidence. Therefore, combining the extracted OpenFace~\cite[][]{baltrusaitis2018openface} features with others is necessary.

\section{Ensemble Bagging}\label{sec:bagging}
To better benchmark the federated learning algorithms, we implemented ensemble learning in the experiments, namely bootstrap aggregating (bagging) with the same bi-LSTM model as base learner.

We employed 15 base learners and hence re-sampled 15 same-sized subsets from the original datasets. To speed up training, we increased batch size to 128 for bagging. 
During inference, the predictions of base models were aggregated through majority vote. We applied soft vote for majority vote, which means that instead of outputting class labels, each base learner output class probability distribution. The probabilities were then averaged, and the final prediction was the class with the highest average probability. We chose this vote strategy to prevent a draw vote where the number of learners is even and both classes get the same amount of voters. It should also be noticed that the base learners in bagging are normally initialized differently, whereas in federated learning the client models commonly share the same starting point.

Ensemble bagging is similar to federated learning in the sense that they both leverage multiple models and apply model aggregation in order to improve model performance over a single model. A key difference between them in this regard lies in that compared to majority vote in bagging, the model averaging in federated learning may take more advantage of model personalization~\cite{tan2022towards, mansour2020three, zhang2023fedala}. We gave a toy example of this effect in the following. The purpose of this example is not to serve as strict proof but as a heuristic.

\subsection{Toy Example}
Consider a binary classification task in deep learning. The task is solved both in bagging with 100 base learners, and in federated learning with 100 clients. Both bagging and federated learning leverage the same model architecture, with a single neuron with sigmoid activation function $sigmoid(x) = \frac{1}{1 + exp(-x)}$ as the last layer in order to output class probability distribution (probability that input sample belongs to class 1). The inverse function of sigmoid is logit function with $logit(x) = ln(\frac{x}{1 - x})$. For bagging, the majority vote follows soft vote strategy, meaning that the final prediction is computed by averaging the class distributions (continuous value between 0.0 and 1.0) rather than class labels (discrete value, either 0 or 1) of all base learners. For federated learning, the vanilla federated aggregation method, namely FedAvg~\cite{mcmahan2017communication}, which simply averages parameters of client models, is adopted.

A new instance belonging to class 1 is presented to the model. Suppose this new sample appears exclusively in one resampled subset in bagging and in the local dataset of a single client in federated learning. As a result, only one specific base learner and one specific client model, respectively, can be trained on this sample. Further, assume currently the aggregated prediction for this sample is 0.49 (class 0) both in bagging and in federated learning. To correct the prediction for this sample, the aggregated output must increase by a minimum of 0.01 to exceed the classification threshold (e.g., 0.5), resulting in a predicted label of class 1. In the case of bagging with 100 base learners, this would require increasing the output of that specific single base model by at least 1.0, which is mathematically infeasible, since each model's output is a probability bounded between 0.0 and 1.0. In contrast, within the federated learning setting, correcting the prediction would require increasing the logit value (the output of the last neuron before sigmoid activation) by approximately 0.04 ($logit(0.5) - logit(0.49) \approx 0.04$). This adjustment can be achieved through joint optimization of the weights and biases of the corresponding client model, or even by simply increasing the bias term of the model’s final layer by 4.0 in the presence of 100 client models, both of which are mathematically feasible during training. In other words, model averaging in federated learning could potentially benefit more from model personalization, whereas in bagging, the effect of model personalization may be overwhelmed by ``wisdom of crowds''.

\section{Hyperparameters} \label{chap:hparam}
\begin{table*}[ht]
    \centering
    \scriptsize
    \caption{Learning rates used in the experiments on the  EngageNet~\cite[][]{singh2023have}, Colorado~\cite[][]{bosch2019automatic}, Korea~\cite[][]{lee2022predicting} and DAiSEE~\cite[][]{gupta2016daisee, kamath2016crowdsourced} datasets. All federated learning methods share the same client learning rates as FedAvg~\cite[][]{mcmahan2017communication}. For FedAvg~\cite[][]{mcmahan2017communication}, FedProx~\cite[][]{li2020federated}, and MOON~\cite[][]{li2021model}, client learning rate is listed in the table, while for FedAdam~\cite[][]{reddi2020adaptive}, FedAwS~\cite[][]{yu2020federated}, and TurboSVM~\cite[][]{wang2024turbosvmfl}, server learning rate is given. All federated learning methods train bi-LSTM, trained on EmoNet~\cite[][]{toisoul2021estimation} and OpenFace gaze~\cite[][]{baltrusaitis2018openface} feature set.}
    \begin{tabular}{c l c c c c c}
    \toprule
        Features & Algorithm & Colorado & Korea & Germany & EngageNet & DAiSEE  \\
        \midrule
        \multirow{7}{*}{ \begin{sideways} \thead{EmoNet + \\ OpenFace gaze}\end{sideways}} & 
         non-FL (bi-LSTM) & $1e-4$ & $1e-2$ & $1e-2.5$ & $1e-4$ & $1e-4$ \\
         & non-FL (MeGlass) & $1e-5.5$ & $1e-5$ & $1e-7.5$ & $1e-6$ & $1e-5.5$ \\
         & non-FL bagging & $1e-2$ & $1e-1$ & $1e-1.5$ & $1e-2$ & $1e-2$ \\
        & FedAvg & $1e-3$ & $1e-2$ & $1e-1$ & $1e-3.5$ & $1e-3$ \\
        & FedAdam & $1e-4.5$ & $1e-3.5$ & $1e-4$ & $1e-4.5$ & $1e-5$ \\
        & FedAwS & $1e-3.5$ & $1e-2$ & $1e-2$ & $1e-3.5$ & $1e-5$ \\
        & FedProx & $1e-3$ & $1e-2$ & $1e-1$ & $1e-3.5$ & $1e-3$ \\
        & MOON & $1e-3$ & $1e-2$ & $1e-1$ & $1e-3.5$ & $1e-3$ \\
        & TurboSVM-FL & $1e-3$ & $1e-3.5$ & $1e-2$ & $1e-3$ & $1e-5$ \\
        \bottomrule
    \end{tabular}
    
    \label{tab:lr}
\end{table*}
\begin{table*}[ht]
    \centering
    \scriptsize
    \caption{Learning rates used in the ablation study on different feature sets of EmoNet~\cite[][]{toisoul2021estimation} and OpenFace~\cite[][]{baltrusaitis2018openface}. The learning rates are obtained when training bi-LSTM on centralized and decentralized (FedAvg~\cite[][]{mcmahan2017communication}) settings on the EngageNet~\cite[][]{singh2023have}, Colorado~\cite[][]{bosch2019automatic}, Korea~\cite[][]{lee2022predicting} and DAiSEE~\cite[][]{gupta2016daisee, kamath2016crowdsourced} datasets.}
    \begin{tabular}{c c c c c c c}
    \toprule
        Features & Algorithm & Colorado & Korea & Germany & EngageNet & DAiSEE  \\
        \midrule
        \multirow{2}{*}{EmoNet}&  non-FL &  $1e-4$ &  $1e-2$ & $1e-2.5$ & $1e-4$ &  $1e-4$ \\
         &  FedAvg &  $1e-3$ &  $1e-1.5$ & $1e-1.5$ & $1e-3.5$ &  $1e-3$ \\
         \midrule
         \multirow{2}{*}{OpenFace} &  non-FL &  $1e-3$ &  $1e-3$ & $1e-4$ & $1e-4$ &  $1e-4$ \\
         &  FedAvg &  $1e-2.5$ &  $1e-3$ & $1e-3.5$ & $1e-3.5$ &  $1e-3$ \\
         \midrule
         \multirow{2}{*}{OpenFace gaze} &  non-FL &  $1e-2$ &  $1e-1$ & $1e-2$ & $1e-1$ &  $1e-2$ \\
         &  FedAvg &  $1e-1$ &  $1e-1$ & $1e-1.5$ & $1e-1$ &  $1e-2$ \\
         \midrule
        \multirow{2}{*}{  \makecell{ EmoNet + \\OpenFace}} &  non-FL &  $1e-4$ &  $1e-2$ & $1e-3.5$ & $1e-4$ &  $1e-4$ \\
         &  FedAvg &  $1e-3$ &  $1e-3$ & $1e-1$ & $1e-3.5$ &  $1e-2.5$ \\
        \bottomrule
    \end{tabular}
    
    \label{tab:ablation_lr}
\end{table*}

The training data was divided into five folds for cross-validation. We conducted a grid search in range $\{10^{-5.5}, \allowbreak10^{-5}, \allowbreak10^{-4.5},10^{-4}, \allowbreak10^{-3.5}, \allowbreak10^{-3}, \allowbreak10^{-2.5}, \allowbreak10^{-2}, \allowbreak10^{-1.5}, \allowbreak10^{-1}\}$ for optimal learning rates with the help of cross-validation. When searching for learning rates for federated learning methods, we first looked for the optimal client learning rates for FedAvg~\cite[][]{mcmahan2017communication} and then applied this learning rate for all other methods. For federated learning algorithms that require a server optimizer, such as FedAdam, FedAwS, and TurboSVM-FL, we carried out a grid search for server learning rates in the same range. The mini-batch size was 4 throughout the experiments, and any user with less than 4 samples was left out for the experiments. The number of client local training epochs was set to 8. For centralized learning, the optimizer was set to stochastic gradient descent (SGD). For federated learning, we chose SGD as the client optimizer and Adam as the server optimizer, as suggested in previous works~\cite[][]{reddi2020adaptive, wang2021field}. A further common assumption in federated learning is that not all clients can participate in every global aggregation round. In this regard, we assumed 50\% of clients attended each aggregation round, and we randomly sampled these clients in each round. The best-performing learning rates are given in Table~\ref{tab:lr}.

While running 5-fold cross-validation on the training set, for the Colorado~\cite[][]{bosch2019automatic} and Korea~\cite[][]{lee2022predicting} datasets, it can happen that all users in some fold share the same single class label regarding glass wearing behavior. Therefore, we manually exchanged a small portion of samples across folds. Table~\ref{tab:dataset_split_details} shows a thorough analysis of the train and test sets for all five datasets. The exact client division of each dataset can be found in our GitHub repository.

\section{Additional Metrics for Federated Learning}
In our case, where the datsets are highly unbalanced and the detection of both classes are equally important, we also reported the weighted $F_1$ score. Instead of focusing exclusively on the minority class (e.g., mind-wandering detection), we aim to evaluate overall model performance across all classes while still ensuring that the minority class contributes meaningfully to the final metric. The weighted $F_1$ score achieves this by computing $F_1$ scores per class and weighting them based on class prevalence, ensuring that both majority and minority classes are appropriately considered, reflecting real-world class distributions. This is particularly important in our setting, as our federated learning approaches must generalize across different datasets where the degree of class imbalance varies. Additionally, misclassification in both classes affects downstream applications, making a balanced evaluation crucial.

The necessity of using the weighted $F_1$ score arises from the practical challenges associated with false detections. In highly imbalanced datasets, the weighted $F_1$ score provides a more reliable measure of model performance than the binary $F_1$ score, as it accounts for class imbalance by averaging metrics weighted by class support. This strategy reflects the actual model performance more accurately, especially when false detections could potentially disrupt the learning process. While real-time detection of disengagement could enhance learning efficiency and improve learning outcomes, any incorrect detection risks disrupting the learning flow of students, potentially undermining the overall learning results. Therefore, it is necessary to determine a corresponding confidence interval for each prediction and only get back to the user with a warning message or a practical exercise when the network has high confidence in predicting mind wandering, disengagement, or boredom. Should warning messages appear too often, especially during periods of active engagement, they might distract learners to the extent that they lose focus on the instructional video.  

\begin{table*}[t]
    \centering
    \scriptsize
    \caption{Federated learning results on the five datasets (Colorado~\cite[][]{bosch2019automatic}, Korea~\cite[][]{lee2022predicting}, Germany~\cite[][]{buhler2024detecting}, EngageNet~\cite[][]{singh2023have}, and DAiSEE~\cite[][]{gupta2016daisee, kamath2016crowdsourced}).  All federated learning results are obtained with bi-LSTM model. The $F_1$ score is reported with the ``weighted'' setting.}
    \begin{tabular}{p{16mm} p{7mm} c c c c c}
    \toprule
        \multicolumn{2}{c}{Performance} & Colorado & Korea & Germany & EngageNet & DAiSEE \\
        \midrule
         non-FL & $F_1$[\%] & 35.7$\pm$13.5 & 75.1$\pm$9.5 & 52.7$\pm$6.3 & 61.5$\pm$1.8 & 54.9$\pm$6.9 \\
         bi-LSTM & $Acc$[\%] & 39.9$\pm$9.3 & 67.1$\pm$13.0 & 52.5$\pm$6.2 & 63.3$\pm$2.9 & 48.2$\pm$6.4\\
         & $AUC$[\%] & 50.8$\pm$7.7 & 75.3$\pm$8.2 & 53.0$\pm$5.4 & 70.0$\pm$1.6 & 54.4$\pm$0.4 \\
         \midrule
         non-FL & $F_1$[\%] & 40.9$\pm$11.9 & 89.9$\pm$3.5 & 54.3$\pm$4.2 & 64.2$\pm$2.7 & 61.6$\pm$1.4 \\
         bagging & $Acc$[\%] & 44.1$\pm$8.3 & 89.5$\pm$5.9 & 55.7$\pm$4.7 & 67.4$\pm$4.7 & 54.9$\pm$1.6 \\
         bootstrap & $AUC$[\%] & 63.0$\pm$1.8 & 74.3$\pm$4.4 & 49.2$\pm$2.7 & 70.0$\pm$0.9 & 55.3$\pm$0.4 \\
         \midrule
        \multirow{2}{*}{FedAvg} & $F_1$[\%] & 64.9$\pm$3.9 & 76.4$\pm$7.1 & 43.6$\pm$13.8 & 64.1$\pm$2.8 & 51.1$\pm$7.3 \\
        & $Acc$[\%] & 63.8$\pm$4.5 & 68.6$\pm$10.0 & 48.4$\pm$7.2 & 63.8$\pm$2.9 & 44.7$\pm$6.8 \\
        & $AUC$[\%] & 63.5$\pm$5.0 & 70.2$\pm$2.0 & 53.3$\pm$2.0 & 69.4$\pm$3.1 & 54.9$\pm$2.5 \\
        \midrule
        \multirow{2}{*}{FedAdam} & $F_1$[\%] & 64.8$\pm$1.6 & 77.1$\pm$10.7 & 53.7$\pm$2.9 & 65.0$\pm$3.0 & 38.0$\pm$19.0 \\
        & $Acc$[\%] & 64.7$\pm$2.6 & 70.5$\pm$13.5 & 53.8$\pm$2.6 & 66.2$\pm$3.3 & 35.7$\pm$14.1 \\
        & $AUC$[\%] & 60.2$\pm$4.7 & 67.6$\pm$10.9 & 54.3$\pm$3.8 & 68.8$\pm$4.3 & 56.1$\pm$1.4 \\
        \midrule
        \multirow{2}{*}{FedAwS} & $F_1$[\%] & 62.7$\pm$2.5 & 78.6$\pm$7.2 & 43.6$\pm$10.1 & 63.3$\pm$2.7 & 52.3$\pm$7.1 \\
        & $Acc$[\%] & 61.3$\pm$2.8 & 71.8$\pm$10.3 & 48.1$\pm$5.1 & 63.7$\pm$3.0 & 45.7$\pm$6.3 \\
        & $AUC$[\%] & 63.3$\pm$4.1 & 70.9$\pm$5.0 & 56.2$\pm$4.3 & 70.7$\pm$2.4 & 54.1$\pm$2.0 \\
        \midrule
        \multirow{2}{*}{FedProx} & $F_1$[\%] & 59.2$\pm$11.0 & 71.8$\pm$9.0 & 40.5$\pm$12.2 & 66.0$\pm$3.6 & 49.3$\pm$11.5 \\
        & $Acc$[\%] & 59.7$\pm$10.3 & 62.8$\pm$11.0 & 47.1$\pm$8.0 & 66.0$\pm$3.7 & 43.5$\pm$9.4 \\
        & $AUC$[\%] & 59.6$\pm$2.7 & 73.7$\pm$6.4 & 53.4$\pm$6.9 & 70.3$\pm$3.5 & 55.5$\pm$1.5 \\
        \midrule
        \multirow{2}{*}{MOON} & $F_1$[\%] & 59.1$\pm$4.8 & 67.7$\pm$18.4 & 49.6$\pm$10.0 & 59.5$\pm$5.7 & 19.9$\pm$10.2\\
        & $Acc$[\%] & 59.3$\pm$7.2 & 59.8$\pm$20.2 & 51.8$\pm$8.8 & 60.2$\pm$4.6 & 23.0$\pm$6.0 \\
        & $AUC$[\%] & 53.5$\pm$6.5 & 66.2$\pm$5.7 & 56.0$\pm$6.9 & 68.0$\pm$4.6 & 55.9$\pm$2.2 \\
        \midrule
        \multirow{2}{*}{TurboSVM} & $F_1$[\%] & 57.7$\pm$9.1 & 63.9$\pm$12.2 & 47.3$\pm$16.4 & 63.4$\pm$3.3 & 49,4$\pm$5.7\\
        & $Acc$[\%] & 56.8$\pm$9.0 & 53.9$\pm$15.1 & 52.4$\pm$8.4 & 64.2$\pm$3.1 & 42.9$\pm$5.0 \\
        & $AUC$[\%] & 60.2$\pm$3.6 & 63.6$\pm$11.9 & 55.6$\pm$2.4 & 66.6$\pm$3.5 & 52.6$\pm$3.2 \\
        \bottomrule
    \end{tabular}
    
    \label{tab:FL_SUM_additional}
\end{table*}

\end{appendices}


\end{document}